\PassOptionsToPackage{unicode}{hyperref}
\PassOptionsToPackage{hyphens}{url}
\PassOptionsToPackage{dvipsnames,svgnames,x11names}{xcolor}

\documentclass[
  letterpaper,
]{article}
\usepackage{xcolor}
\usepackage{amsmath,amssymb}
\usepackage{iftex}
\usepackage[T1]{fontenc}
\usepackage[utf8]{inputenc}
\usepackage{textcomp}
\usepackage{lmodern}
\IfFileExists{upquote.sty}{\usepackage{upquote}}{}
\IfFileExists{microtype.sty}{
  \usepackage[]{microtype}
  \UseMicrotypeSet[protrusion]{basicmath}
}{}
\makeatletter
\@ifundefined{KOMAClassName}{
  \IfFileExists{parskip.sty}{
    \usepackage{parskip}
    \setlength{\parindent}{0pt}
    \setlength{\parskip}{6pt plus 2pt minus 1pt}}
}{
  \KOMAoptions{parskip=half}}
\makeatother

\usepackage{color}
\usepackage{fancyvrb}

\DefineVerbatimEnvironment{Highlighting}{Verbatim}{commandchars=\\\{\}}

\usepackage{framed}
\definecolor{shadecolor}{RGB}{248,248,248}
\newenvironment{Shaded}{\begin{snugshade}}{\end{snugshade}}

\newcommand{\AttributeTok}[1]{\textcolor[rgb]{0.13,0.29,0.53}{#1}}

\newcommand{\CommentTok}[1]{\textcolor[rgb]{0.56,0.35,0.01}{\textit{#1}}}

\newcommand{\ConstantTok}[1]{\textcolor[rgb]{0.56,0.35,0.01}{#1}}

\newcommand{\DecValTok}[1]{\textcolor[rgb]{0.00,0.00,0.81}{#1}}

\newcommand{\FunctionTok}[1]{\textcolor[rgb]{0.13,0.29,0.53}{\textbf{#1}}}

\newcommand{\NormalTok}[1]{#1}

\newcommand{\OtherTok}[1]{\textcolor[rgb]{0.56,0.35,0.01}{#1}}

\newcommand{\SpecialCharTok}[1]{\textcolor[rgb]{0.81,0.36,0.00}{\textbf{#1}}}

\newcommand{\StringTok}[1]{\textcolor[rgb]{0.31,0.60,0.02}{#1}}

\usepackage{longtable,booktabs,array}
\usepackage{calc}

\usepackage{etoolbox}
\makeatletter
\patchcmd\longtable{\par}{\if@noskipsec\mbox{}\fi\par}{}{}
\makeatother

\IfFileExists{footnotehyper.sty}{\usepackage{footnotehyper}}{\usepackage{footnote}}
\makesavenoteenv{longtable}
\usepackage{graphicx}
\makeatletter
\newsavebox\pandoc@box
\newcommand*\pandocbounded[1]{
  \sbox\pandoc@box{#1}%
  \Gscale@div\@tempa{\textheight}{\dimexpr\ht\pandoc@box+\dp\pandoc@box\relax}%
  \Gscale@div\@tempb{\linewidth}{\wd\pandoc@box}%
  \ifdim\@tempb\p@<\@tempa\p@\let\@tempa\@tempb\fi%
  \ifdim\@tempa\p@<\p@\scalebox{\@tempa}{\usebox\pandoc@box}%
  \else\usebox{\pandoc@box}%
  \fi%
}

\def\fps@figure{htbp}
\makeatother

\NewDocumentCommand\citeproctext{}{}
\NewDocumentCommand\citeproc{mm}{%
  \begingroup\def\citeproctext{#2}\cite{#1}\endgroup}
\makeatletter
 \let\@cite@ofmt\@firstofone
 \def\@biblabel#1{}
 \def\@cite#1#2{{#1\if@tempswa , #2\fi}}
\makeatother
\newlength{\cslhangindent}
\newlength{\csllabelwidth}
\newenvironment{CSLReferences}[2]
 {\begin{list}{}{%
  \setlength{\itemindent}{0pt}
  \setlength{\leftmargin}{0pt}
  \setlength{\parsep}{0pt}
  \ifodd #1
   \setlength{\leftmargin}{\cslhangindent}
   \setlength{\itemindent}{-1\cslhangindent}
  \fi
  \setlength{\itemsep}{#2\baselineskip}}}
 {\end{list}}
\usepackage{calc}

\providecommand{\tightlist}{%
  \setlength{\itemsep}{0pt}\setlength{\parskip}{0pt}}

\usepackage[T1]{fontenc}
\usepackage{newtxtext}
\usepackage{newtxmath}
\usepackage[scaled=0.95]{inconsolata}
\usepackage{microtype}
\usepackage[letterpaper,margin=1in]{geometry}
\usepackage{caption}
\usepackage{float}

\usepackage{hyperref}
\hypersetup{
  colorlinks=true,
  linkcolor=black,
  citecolor=black,
  urlcolor=black,
  pdfauthor={},
  pdftitle={}
}
\usepackage[capitalize,nameinlink]{cleveref}

\usepackage{fvextra}
\usepackage{seqsplit}
\usepackage{xcolor}
\usepackage[most]{tcolorbox}
\tcbuselibrary{breakable}

\definecolor{codebg}{HTML}{F9F9FB}
\definecolor{codeframe}{HTML}{D0D0D5}
\definecolor{keyword}{HTML}{0044AA}
\definecolor{string}{HTML}{006600}
\definecolor{comment}{HTML}{7A7A7A}

\IfFileExists{upquote.sty}{\usepackage{upquote}}{}

\renewenvironment{Shaded}{%
  \begin{tcolorbox}[
    enhanced,
    breakable,
    colback=codebg,
    colframe=black!40,
    boxrule=0.8pt,
    left=6pt, right=6pt, top=6pt, bottom=6pt,
    arc=4pt,
    outer arc=4pt,
    before skip=8pt, after skip=10pt
  ]%
}{%
  \end{tcolorbox}%
}

\makeatletter
\newcommand{\addverbbreaks}{%
  \catcode`\/=\active
  \lccode`\~=`\/
  \lowercase{\def~}{/\penalty0\hskip0pt\relax}%
  \catcode`\_=\active
  \lccode`\~=`\_
  \lowercase{\def~}{\_\penalty0\hskip0pt\relax}%
  \catcode`\-=\active
  \lccode`\~=`\-
  \lowercase{\def~}{-\penalty0\hskip0pt\relax}%
  \catcode`\.=\active
  \lccode`\~=`\.
  \lowercase{\def~}{.\penalty0\hskip0pt\relax}%
}
\makeatother

\RecustomVerbatimEnvironment{Highlighting}{Verbatim}{
  breaklines=true,
  breakanywhere=true,
  breakindent=1em,
  breaksymbolleft=,
  commandchars=\\\{\},
  fontsize=\small,
  formatcom=\addverbbreaks
}

\RecustomVerbatimEnvironment{verbatim}{Verbatim}{
  breaklines=true,
  breakanywhere=true,
  breakindent=1em,
  breaksymbolleft=,
  fontsize=\small,
  formatcom=\addverbbreaks
}

\makeatletter
\renewcommand{\Verb}{\Verb[formatcom=\color{black!85},fontsize=\small]}
\makeatother

\usepackage{titlesec}
\titleformat{\section}{\bfseries\large}{\thesection}{1em}{}
\titleformat{\subsection}{\bfseries\normalsize}{\thesubsection}{1em}{}
\titlespacing*{\section}{0pt}{1.2em plus .2em minus .1em}{0.5em}
\titlespacing*{\subsection}{0pt}{1em plus .2em minus .1em}{0.4em}

\providecommand{\pkg}[1]{\texttt{#1}}
\providecommand{\proglang}[1]{\textsf{#1}}

\usepackage{enumitem}
\setlist{nosep}

\DeclareUnicodeCharacter{0007}{}

\newenvironment{qsnl}{%
  \begingroup
  \small
  \setlength{\parindent}{0pt}%
  \setlength{\parskip}{0.5\baselineskip}%
}{%
  \endgroup
}

\usepackage{xcolor}
\usepackage{tikz}

\newcommand{\feBadgeSize}{2.5} %

\newcommand{\feBadge}[5][white]{%
  \tikz[baseline=(n.base)]\node[
    circle,
    inner sep=0pt,
    minimum size=#5ex,
    fill=#2,
    text=#1,
    font=\sffamily\bfseries\ifnum#4>9\tiny\else\footnotesize\fi
  ](n){#3#4};%
}

\newcommand{\Sbadge}[2][teal!60!black]{%
  \feBadge[white]{#1}{S}{#2}{\feBadgeSize}%
}
\newcommand{\Dbadge}[2][purple!70!black]{%
  \feBadge[white]{#1}{D}{#2}{\feBadgeSize}%
}

  \definecolor{S1color}{HTML}{1E4F8A}
  \definecolor{S2color}{HTML}{245A98}
  \definecolor{S3color}{HTML}{2B66A6}
  \definecolor{S4color}{HTML}{3271B4}
  \definecolor{S5color}{HTML}{3A7CC1}
  \definecolor{S6color}{HTML}{4488CD}
  \definecolor{S7color}{HTML}{4E94D9}

  \definecolor{D1color}{HTML}{7A5A00}
  \definecolor{D2color}{HTML}{846100}
  \definecolor{D3color}{HTML}{8E6800}
  \definecolor{D4color}{HTML}{986F00}
  \definecolor{D5color}{HTML}{A17600}
  \definecolor{D6color}{HTML}{AB7D00}
  \definecolor{D7color}{HTML}{B58400}
  \definecolor{D8color}{HTML}{BF8B00}
  \definecolor{D9color}{HTML}{C89200}
  \definecolor{D10color}{HTML}{D19900}
  \definecolor{D11color}{HTML}{DAA000}
  \definecolor{D12color}{HTML}{E3A700}
\usepackage{booktabs}
\usepackage{caption}
\usepackage{longtable}
\usepackage{colortbl}
\usepackage{array}
\usepackage{anyfontsize}
\usepackage{multirow}
\makeatletter
\@ifpackageloaded{caption}{}{\usepackage{caption}}
\AtBeginDocument{%
\ifdefined\contentsname
  \renewcommand*\contentsname{Table of contents}
\else
  \newcommand\contentsname{Table of contents}
\fi
\ifdefined\listfigurename
  \renewcommand*\listfigurename{List of Figures}
\else
  \newcommand\listfigurename{List of Figures}
\fi
\ifdefined\listtablename
  \renewcommand*\listtablename{List of Tables}
\else
  \newcommand\listtablename{List of Tables}
\fi
\ifdefined\figurename
  \renewcommand*\figurename{Figure}
\else
  \newcommand\figurename{Figure}
\fi
\ifdefined\tablename
  \renewcommand*\tablename{Table}
\else
  \newcommand\tablename{Table}
\fi
}
\@ifpackageloaded{float}{}{\usepackage{float}}
\floatstyle{ruled}
\@ifundefined{c@chapter}{\newfloat{codelisting}{h}{lop}}{\newfloat{codelisting}{h}{lop}[chapter]}
\floatname{codelisting}{Listing}

\makeatother
\makeatletter
\@ifpackageloaded{caption}{}{\usepackage{caption}}
\@ifpackageloaded{subcaption}{}{\usepackage{subcaption}}
\makeatother
\usepackage{bookmark}
\IfFileExists{xurl.sty}{\usepackage{xurl}}{} 
\hypersetup{
  pdftitle={flowengineR: A Modular and Extensible Framework for Fair and Reproducible Workflow Design in R},
  pdfkeywords={R, workflow, algorithmic fairness, engine, FAIR
principles, distributed statistical computing},
  colorlinks=true,
  linkcolor={black},
  filecolor={black},
  citecolor={black},
  urlcolor={black},
  pdfcreator={LaTeX via pandoc}}

\begin{document}


\begin{center}
  {\LARGE flowengineR: A Modular and Extensible Framework for Fair and
Reproducible Workflow Design in R \par}
  \vspace{1em}

  {\large
      \mbox{%
              \href{https://orcid.org/0009-0003-7054-1937}{Maximilian
Willer}%
      \textsuperscript{%
        1, 2
      }%
      \textsuperscript{*}%
    }%
    \nobreak\hspace{0pt},\thinspace     \mbox{%
              \href{https://orcid.org/0000-0001-7815-4809}{Peter
Ruckdeschel}%
      \textsuperscript{%
        1
      }%
    }%
      \par}

  \vspace{0.8em}

  {\footnotesize
      \noindent [1] Institute for Mathematics, School of Mathematics and
Sciences,%
    \\ Carl von Ossietzky University Oldenburg%
    \\ Ammerländer Heerstraße 114--118, 26129 Oldenburg,
Germany\par\vspace{0.8em}
      \noindent [2] While writing employed at HASPA Finanzholding%
    \\ Dammtorstraße 1, 20354 Hamburg, Germany\par\vspace{0.8em}
    \par \textsuperscript{*} Corresponding author:
maximilian.willer@uol.de
  }
\end{center}

\vspace{1.0em}

  \begin{abstract}
  \vspace{1.0em}

  \noindent \pkg{flowengineR} is an \proglang{R} package designed to
  provide a modular and extensible framework for building reproducible
  algorithmic workflows for general-purpose machine learning pipelines.

  \noindent It is motivated by the rapidly evolving field of algorithmic
  fairness --- the PhD topic of the first author --- where new metrics,
  mitigation strategies, and machine learning methods continuously
  emerge. A central challenge in fairness, but also far beyond, is that
  existing toolkits either focus narrowly on single interventions or
  treat reproducibility and extensibility as secondary considerations
  rather than core design principles.

  \noindent \pkg{flowengineR} addresses this by introducing a unified
  architecture of standardized \emph{engines} for data splitting,
  execution, preprocessing, training, inprocessing, postprocessing,
  evaluation, and reporting. Each engine encapsulates one methodological
  task yet communicates via a lightweight interface, ensuring workflows
  remain transparent, auditable, and easily extensible.

  \noindent Although implemented in \proglang{R}, \pkg{flowengineR}
  builds on ideas from workflow languages (\texttt{CWL}
  (\citeproc{ref-piccoloSimplifyingDevelopmentPortable2021}{Piccolo et
  al. 2021}), \texttt{YAWL}
  (\citeproc{ref-vanderaalstYAWLAnotherWorkflow2005}{Van Der Aalst and
  Ter Hofstede 2005})), graph-oriented visual programming languages
  (\texttt{KNIME}), and \proglang{R} frameworks (\pkg{BatchJobs},
  \pkg{batchtools}). Its emphasis, however, is less on orchestrating
  engines for resilient parallel execution but rather on the
  straightforward setup and management of distinct engines as data
  structures. This orthogonalization enables distributed
  responsibilities, independent development, and streamlined
  integration.

  \noindent In the context of fairness, by structuring fairness methods
  as interchangeable engines, \pkg{flowengineR} lets researchers
  integrate, compare, and evaluate interventions across the modeling
  pipeline. At the same time, the architecture generalizes to
  explainability, robustness, and compliance metrics without core
  modifications. While motivated by fairness, flowengineR ultimately
  provides a general infrastructure for any workflow context where
  reproducibility, transparency, and extensibility are essential.

  \noindent The source code and all materials to reproduce this paper
  are available at: \url{https://github.com/mwiller1991/flowengineR}
  \end{abstract}

\begin{center}
  \begin{qsnl}
    \emph{This article is part of the cumulative doctoral thesis of
Maximilian Willer at the Carl von Ossietzky University Oldenburg.}
  \end{qsnl}
\end{center}

\vspace{0.5\baselineskip}



\section{\texorpdfstring{Introduction }{Introduction }}\label{sec-1}

\subsection{\texorpdfstring{Motivation \& Problem Statement
}{Motivation \& Problem Statement }}\label{motivation-problem-statement}

Modern data analysis is rarely a sequence of isolated steps. Instead, it
is increasingly structured as a pipeline, where multiple tasks are
combined into a coherent process. In node-based workflow environments
such as KNIME, workflows are represented as interconnected processing
nodes whose input and output ports define the flow of data between steps
(\citeproc{ref-knimeDocumentation2025}{KNIME AG 2025}). A common
distinction is made between orchestration, where a central conductor
directs execution, and choreography, where interaction follows
distributed protocols (this is the prevalent communication strategy,
e.g., in Bischl et al.
(\citeproc{ref-bischlBatchJobsBatchExperimentsAbstraction2015}{2015})).
In this paper we introduce the \proglang{R} package
\textbf{\pkg{flowengineR}} which is situated in between these two
extremes. While orchestration coordinates the overall workflow,
individual components are not expected to communicate directly. Instead,
the framework emphasizes the nodes themselves: modular building blocks
that encapsulate abstract tasks such as splitting, training, or
reporting. These nodes---called \emph{engines}---are given semantic
labels that make them transparent and exchangeable (independently from
the other nodes). Data exchange between engines is handled through
lightweight list structures, which remain deliberately informal yet
effective (as implemented in Landau
(\citeproc{ref-landauTargetsPackageDynamic2021}{2021})). This design
ensures that pipelines remain flexible and auditable without introducing
rigid dependencies.

Our interest in workflow design is motivated by questions of algorithmic
fairness, which will therefore serve as a running example in this paper.
Fairness is particularly instructive because interventions can take
place at different stages of a pipeline and require access to diverse
types of information. Yet the infrastructure provided by flowengineR is
not confined to fairness alone; similar requirements arise in domains
such as explainability, robustness, and privacy. This broader
applicability motivates the architectural perspective developed in the
remainder of the paper.

The implications of fairness for pipeline design are substantial. Unlike
conventional performance evaluation, fairness cannot be addressed at a
single point of the workflow. Instead, it may require interventions at
multiple stages: preprocessing methods adjust the data, inprocessing
methods alter the training procedure, and postprocessing methods
recalibrate outputs. In addition, fairness requires dedicated
measurement at the evaluation stage, where group-specific outcomes are
compared to quantify disparities and assess whether mitigation
strategies succeed (\citeproc{ref-mehrabiSurveyBiasFairness2022}{Mehrabi
et al. 2022}). Fairness interventions also differ conceptually, since
different fairness mechanisms rely on different assumptions about what
it means for a decision process to be fair
(\citeproc{ref-friedlerImpossibilityFairnessDifferent2021}{Friedler et
al. 2021}). From a workflow perspective, this translates into
heterogeneous information requirements: preprocessing methods operate on
data distributions, inprocessing methods on optimization objectives,
postprocessing methods on model outputs, and evaluation metrics on
protected-group outcomes. What is missing is a general-purpose
infrastructure that makes fairness a native element of workflow design,
while remaining open to future developments. Without such a framework,
it is difficult to establish a unified and extensible system,
particularly in a methodological field that continues to evolve at high
speed.

This requirement is reinforced by emerging AI governance frameworks,
which increasingly link fairness-related concerns with transparency,
documentation, and risk management. Although such frameworks do not
prescribe a uniform set of statistical fairness metrics, they strengthen
the need for workflow infrastructures that are not only methodologically
flexible but also reproducible, auditable, and traceable. This broader
compliance dimension is revisited in Section~\ref{sec-5.3}.

The methodological landscape in which workflow frameworks operate is
highly dynamic. New definitions of fairness, novel interventions, and
additional evaluation criteria are introduced at a pace that no central
framework can fully anticipate
(\citeproc{ref-mehrabiSurveyBiasFairness2022}{Mehrabi et al. 2022}).
This challenge is not limited to fairness: explainability, robustness,
and compliance are equally evolving domains, each generating fresh
requirements for workflow design. Attempting to maintain a closed or
centrally managed system in such an environment would quickly lead to
obsolescence. What is needed instead is decentralized extensibility: the
ability for users and communities to implement new engines
independently, without altering the core of the framework. By embedding
openness and modularity at the architectural level,
\textbf{\pkg{flowengineR}} addresses the dynamism of the field and
provides a sustainable path for future developments.


\subsection{\texorpdfstring{Our Vision }{Our Vision }}\label{our-vision}

Our vision for \textbf{\pkg{flowengineR}}
(\citeproc{ref-willerRuckdeschelFlowengineR2025}{Willer and Ruckdeschel
2025}) is to represent algorithmic workflows as a chain of modular,
reusable building blocks that we call \emph{engines.} Each engine
encapsulates a specific task---such as data splitting, model training,
or evaluation---while exposing a standardized interface to the rest of
the workflow.

This vision rests on three technical principles: \textbf{modularity},
\textbf{class-free interfaces}, and \textbf{structured input--output
with smart defaults}. These principles are design choices of
flowengineR. They are intended to support transparent and inspectable
workflows, while scalable execution connects conceptually to
batch-oriented R infrastructures such as BatchJobs and batchtools, which
provide abstraction mechanisms for executing computational experiments
in batch environments
(\citeproc{ref-bischlBatchJobsBatchExperimentsAbstraction2015}{Bischl et
al. 2015}). To capture these qualities systematically, we introduce a
set of \textbf{desirable properties} that serve as guiding design
principles for the framework. These desirables include orthogonal
process steps, modularity by design, class-free interfaces, structured
input-output with smart defaults, compatibility with distributed
computing, accessibility for domain experts, and regulatory suitability.
They provide a vocabulary for describing what sustainable workflow
systems should achieve and will be revisited in detail in
Section~\ref{sec-2}.

A central design goal of \pkg{flowengineR} is accessibility for its
target audience: domain experts who may not have extensive training in
software engineering. By relying on familiar \proglang{R} idioms, plain
functions, and structured control objects, the framework is intended to
lower the barrier for researchers in applied fields to implement and
adapt methods. This usability for non-\proglang{R}-experts is
particularly relevant in domains where methodological innovation often
outpaces available tooling, such as economics, social sciences, or risk
modeling. Beyond accessibility, the framework is designed for use in
regulated environments. Workflows created with \pkg{flowengineR} are
fully auditable, reproducible, and traceable, qualities that are
essential in sectors such as banking or healthcare, where transparency
and compliance are mandatory. By combining methodological openness with
compliance-oriented documentation, \pkg{flowengineR} supports
experimentation and innovation while strengthening the reproducibility,
traceability, and interpretability of results in sensitive application
contexts.


\subsection{\texorpdfstring{Theoretical Background \& Related Work
}{Theoretical Background \& Related Work }}\label{theoretical-background-related-work}

Fairness in machine learning and in algorithms in general has been
approached from multiple angles, leading to a broad taxonomy of
mitigation strategies. A common classification distinguishes between
\textbf{preprocessing}, \textbf{inprocessing}, and
\textbf{postprocessing} interventions
(\citeproc{ref-mehrabiSurveyBiasFairness2022}{Mehrabi et al. 2022}). The
concrete methodological details of these intervention types are not the
focus of the present article and will therefore only be revisited where
relevant in the later use-case discussion.

To position \textbf{\pkg{flowengineR}} within this heterogeneous
landscape, it is useful to distinguish between frameworks that operate
natively within \proglang{R} and those that have been developed outside
the \proglang{R} ecosystem. The former highlight the strengths and
limitations of solutions embedded in the statistical computing
environment, while the latter showcase more general-purpose
infrastructures that have gained traction in applied machine learning
and software engineering. This distinction structures the following
discussion and is not merely technical. Frameworks outside \proglang{R}
often provide mature solutions for orchestration, deployment, graphical
workflow design, and integration across heterogeneous production
environments. Their strength lies in broad interoperability and
operational robustness. Native \proglang{R} implementations, by
contrast, offer different advantages: they keep statistical modeling,
diagnostics, intermediate results, reporting, and documentation within
one coherent programming environment. This enables common data
structures for intermediate outputs, unified diagnostic and logging
objects, and direct integration with reproducible reporting tools such
as \pkg{rmarkdown} or \pkg{Quarto}. For method-oriented researchers and
domain experts, this reduces translation effort across ecosystems and
makes workflows easier to inspect, adapt, and extend.

Outside the \proglang{R} ecosystem, a variety of frameworks provide
powerful but distinct approaches to workflow design and fairness
interventions. General workflow systems such as \emph{KNIME}
(\citeproc{ref-knimeDocumentation2025}{KNIME AG 2025}), \emph{MLflow}
(\citeproc{ref-chenDevelopmentsMLflowSystem2020}{Chen et al. 2020}), and
\emph{Apache Airflow}
(\citeproc{ref-yasminEmpiricalStudyDevelopers2025}{Yasmin et al. 2025})
focus on structuring and orchestrating complex pipelines. KNIME (short
for Konstanz Information Miner) uses graphical nodes to represent
diverse tasks, while MLflow and Airflow emphasize experiment tracking,
deployment, and large-scale execution. These frameworks are widely used
in practice, yet they remain detached from \proglang{R}'s statistical
environment and require translation across ecosystems. In the domain of
fairness, prominent toolkits include \emph{AI Fairness 360}
(\citeproc{ref-bellamyAIFairness3602019}{Bellamy et al. 2019}),
\emph{Fairlearn} (\citeproc{ref-Weertsfairlearn2023}{Weerts et al.
2023}), and \emph{BeFair}
(\citeproc{ref-castelnovoBeFairAddressingFairness2021}{Castelnovo et al.
2020}). These tools provide important collections of fairness metrics
and mitigation methods, ranging from reweighting to relabeling
strategies. However, fairness is typically implemented as a set of
metrics or interventions rather than as a general workflow architecture.
They therefore demonstrate the breadth of available fairness tooling,
but they do not provide a general \proglang{R}-native engine
architecture for end-to-end workflow design.

Within the \proglang{R} ecosystem, several frameworks have been proposed
to address workflow design, reproducibility, and modeling. Packages such
as \emph{\pkg{targets}}
(\citeproc{ref-landauTargetsPackageDynamic2021}{Landau 2021}) and its
predecessor \emph{\pkg{drake}} provide project-level pipeline management
with dependency graphs and caching, while \emph{\pkg{BatchJobs}} and
\emph{\pkg{batchtools}}
(\citeproc{ref-bischlBatchJobsBatchExperimentsAbstraction2015}{Bischl et
al. 2015}; \citeproc{ref-langBatchtoolsToolsWork2017}{Lang et al. 2017})
abstract execution logic for high-performance computing environments.
Complementing these, CRAN Task Views such as Reproducible Research,
Machine Learning, and High-Performance Computing
(\citeproc{ref-cranCRANTaskViews2025}{CRAN 2025}) catalogue a wide range
of packages for documentation, statistical learning, and
parallelization, yet they remain descriptive rather than
infrastructural. Frameworks such as \emph{\pkg{caret}}
(\citeproc{ref-kuhnCaret2008}{Kuhn 2008}), \emph{\pkg{mlr}}
(\citeproc{ref-bischlMlrMachineLearning2013}{Bischl et al. 2013}), and
\emph{\pkg{mlr3}} (\citeproc{ref-langMlr3ModernObjectoriented2019}{Lang
et al. 2019}) demonstrate \proglang{R}'s tradition of modularity, though
they focus primarily on modeling, tuning, and resampling rather than
general workflow integration across preprocessing, training,
postprocessing, evaluation, reporting, and publishing. In terms of
fairness, the \emph{\pkg{fairadapt}} package
(\citeproc{ref-pleckoFairadaptCausalReasoning2024}{Plecko et al. 2024})
illustrates how counterfactual adjustments can be applied at the
preprocessing stage, offering a specialized but conceptually important
approach. The \emph{\pkg{fairness}} package
(\citeproc{ref-kozodoiFairnessAlgorithmicFairness2019}{Kozodoi and V.
Varga 2019}) provides broad metrics and visualization capabilities but
no mitigation methods. Building on the \pkg{mlr3} ecosystem,
\emph{\pkg{mlr3fairness}}
(\citeproc{ref-pfistererMlr3fairnessFairnessAuditing2025}{Pfisterer et
al. 2025}) integrates fairness metrics and interventions directly into
established pipelines, strengthening fairness support in modern
\proglang{R} workflows.

Taken together, existing frameworks demonstrate both the breadth and the
fragmentation of current solutions. Outside \proglang{R}, workflow
systems offer mature orchestration and deployment capabilities, while
fairness toolkits provide important metric and mitigation libraries.
Within \proglang{R}, reproducibility frameworks, modeling ecosystems,
and fairness packages provide strong building blocks, but they are
either focused on execution management, modeling, or specific fairness
interventions. What is still missing is a unifying architecture that
abstracts workflows into modular, interchangeable components while
integrating fairness interventions as native elements. It is precisely
this gap that \pkg{flowengineR} seeks to address by providing a general,
extensible, and \proglang{R}-native solution.


\subsection{\texorpdfstring{Research Gap
}{Research Gap }}\label{sec-1_4}

Existing frameworks provide valuable tools but fall short of offering
the architectural openness and extensibility required for long-term
sustainability. Toolkits such as \emph{Fairlearn}
(\citeproc{ref-Weertsfairlearn2023}{Weerts et al. 2023}) and \emph{AI
Fairness 360} (\citeproc{ref-bellamyAIFairness3602019}{Bellamy et al.
2019}) have popularized fairness metrics and mitigation algorithms,
while domain-specific approaches like \emph{befair}
(\citeproc{ref-castelnovoBeFairAddressingFairness2021}{Castelnovo et al.
2020}) illustrate concrete methods for applications in finance. Yet the
focus of befair is primarily methodological: it offers a curated set of
fairness interventions but does not provide a general workflow
infrastructure. In contrast, \pkg{flowengineR} is designed as a
framework for sustainable extensibility, where infrastructure comes
first and individual methods are realized as interchangeable engines.
More generally, existing fairness toolkits provide important metrics and
mitigation methods, and some explicitly support extension. However,
their extensibility is primarily organized around fairness-specific
methods rather than a general, R-native workflow architecture in which
arbitrary workflow stages can be represented as interchangeable engines.
This mismatch between rapid methodological innovation and rigid
infrastructures reveals a clear research gap: current frameworks lack
the architectural openness and modular extensibility needed to support
sustainable fairness-aware workflows.

To address this gap, we define a set of desirable properties that
articulate what a sustainable workflow system should achieve. These
include \hypertarget{S1}{}\textbf{\Sbadge[S1color]{1} Orthogonal Process
Steps}, which separate concerns and enable distributed responsibility;
\hypertarget{S2}{}\textbf{\Sbadge[S2color]{2} Modularity by Design},
ensuring that each building block is independently replaceable;
\hypertarget{S3}{}\textbf{\Sbadge[S3color]{3} Class-Free Interfaces},
which avoid the rigidity of object-oriented hierarchies and keep
workflows lightweight; \hypertarget{S4}{}\textbf{\Sbadge[S4color]{4}
Structured Input--Output with Smart Defaults}, balancing transparency
with usability; \hypertarget{S5}{}\textbf{\Sbadge[S5color]{5}
Distributed Statistical Computing Compatibility}, aligning with scalable
execution in batch and high-performance computing environments
(\citeproc{ref-bischlBatchJobsBatchExperimentsAbstraction2015}{Bischl et
al. 2015}); \hypertarget{S6}{}\textbf{\Sbadge[S6color]{6} Domain-Expert
Accessibility}, allowing researchers with basic \proglang{R} knowledge
to implement new engines without deep IT expertise; and
\hypertarget{S7}{}\textbf{\Sbadge[S7color]{7} Regulatory Suitability},
ensuring that workflows are reproducible, auditable, and well-suited for
regulated environments such as finance or healthcare.

\begin{figure}

\centering{

\includegraphics[width=1\linewidth,height=\textheight,keepaspectratio]{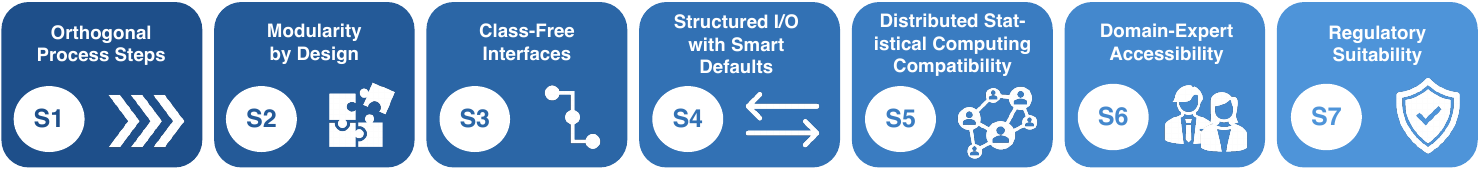}

}

\caption{\label{fig-desirables}\emph{Key desirable properties for
sustainable workflow frameworks. These desirables summarize the research
gap and form the architectural foundation of \pkg{flowengineR}.}}

\end{figure}%

Taken together, these desirables (Figure~\ref{fig-desirables}) define
the architectural foundation of \textbf{\pkg{flowengineR}} and resonate
with well-established design patterns
(\citeproc{ref-gammaDesignPatternsElements1995}{Gamma 1995}) in software
engineering (e.g., Strategy, Template Method, Facade), which will be
operationalized in Section~\ref{sec-2} through concrete design
principles.

\begin{figure}[H]

\centering{

\includegraphics[width=1\linewidth,height=\textheight,keepaspectratio]{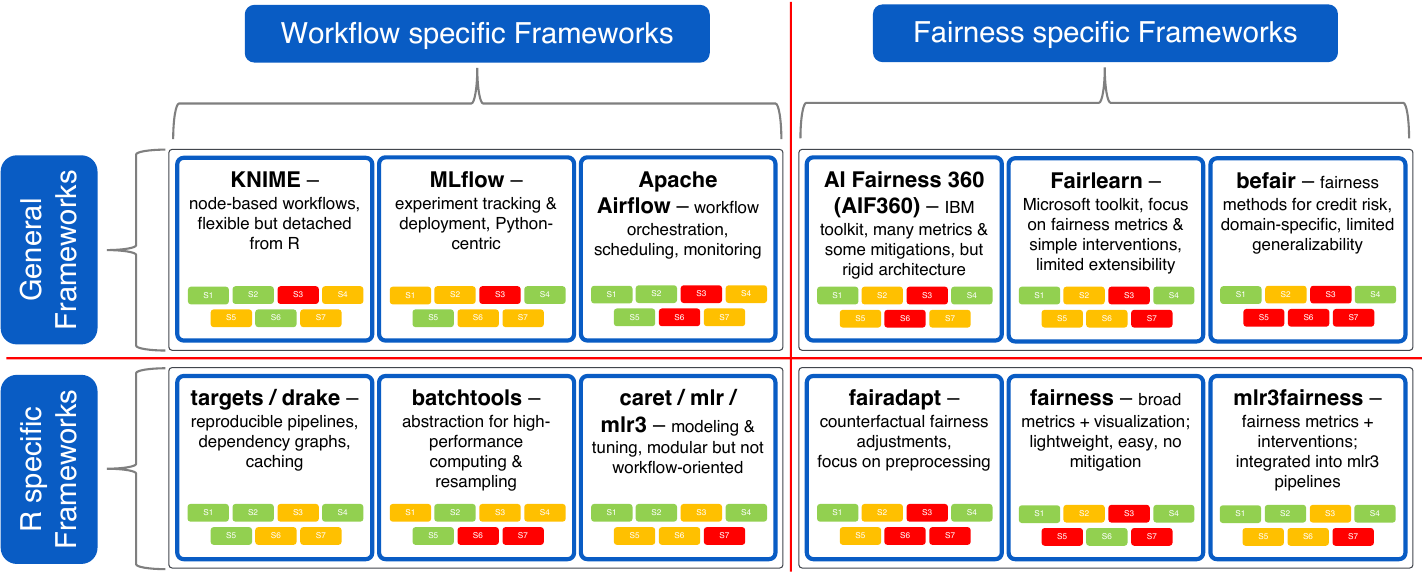}

}

\caption{\label{fig-existingframeworks}\emph{Comparative mapping of
existing workflow and fairness frameworks against the seven desirables
(\Sbadge[S1color]{1}--\Sbadge[S7color]{7}). Legend: Green = property
fulfilled, Yellow = partially fulfilled, Red = not fulfilled}}

\end{figure}

Figure~\ref{fig-existingframeworks} provides a comparative overview of
existing workflow and fairness frameworks, mapped against the desirables
(\hyperlink{S1}{\Sbadge[S1color]{1}}--\hyperlink{S7}{\Sbadge[S7color]{7}}).
It illustrates that while current solutions cover some aspects
partially, none of them jointly realize the full set of properties that
define sustainable, extensible workflow infrastructures. This visual gap
analysis further motivates the development of \pkg{flowengineR} as a
framework explicitly designed to address all seven desirables in an
integrated manner. The mapping in Figure~\ref{fig-existingframeworks} is
an interpretive comparison by the authors. Green, yellow, and red do not
indicate absolute quality judgments, but whether a framework explicitly
provides the corresponding property as a central architectural feature,
partially supports it, or does not target it as part of its core design.
The assessment is based on published documentation and the scope of each
framework.

\subsection{\texorpdfstring{Paper Structure
}{Paper Structure }}\label{paper-structure}

The remainder of this article is organized as follows.
Section~\ref{sec-2} introduces the architectural foundations of
\pkg{flowengineR} and explains how the desirable properties identified
above are translated into concrete design principles.
Section~\ref{sec-3} describes the implementation, including engines,
wrappers, controllers, and extension mechanisms. Section~\ref{sec-4}
presents a fairness-aware modeling use case that demonstrates how
preprocessing, inprocessing, postprocessing, evaluation, and reporting
can be integrated within one workflow. Section~\ref{sec-5} generalizes
the framework to further application domains such as explainability,
robustness, and compliance. Section~\ref{sec-6} summarizes the
contributions, discusses limitations, and outlines future work.
Throughout the paper, we refer to the seven \emph{desirables}
(\hyperlink{S1}{\Sbadge[S1color]{1}}--\hyperlink{S7}{\Sbadge[S7color]{7}})
introduced in Section~\ref{sec-1_4} and the twelve \emph{design
principles}
(\hyperlink{D1}{\Dbadge[D1color]{1}}--\hyperlink{D12}{\Dbadge[D12color]{12}})
operationalized in Section~\ref{sec-2}.


\section{\texorpdfstring{\textbf{Concept \& Architecture of the
Framework }}{Concept \& Architecture of the Framework }}\label{sec-2}

In the following, we use the shorthand notation introduced in
Section~\ref{sec-1_4} and Section~\ref{sec-2_2}: desirables
(\hyperlink{S1}{\Sbadge[S1color]{1}}--\hyperlink{S7}{\Sbadge[S7color]{7}})
denote the overarching architectural goals, while design principles
(\hyperlink{D1}{\Dbadge[D1color]{1}}--\hyperlink{D12}{\Dbadge[D12color]{12}})
represent their operationalization within the framework. These
abbreviations are used consistently across Section 2 to highlight how
individual architectural elements connect back to the identified
research gap.


\subsection{\texorpdfstring{The Control Object Paradigm
}{The Control Object Paradigm }}\label{the-control-object-paradigm}

At the heart of \pkg{flowengineR} lies the concept of the \emph{control
object}, which provides a structured and transparent way of specifying
workflow configurations. Instead of hard-coding individual steps, users
describe their workflow by assembling a set of control parameters that
define which engines are to be used, how they should interact, and what
outputs are to be produced. This abstraction serves a dual purpose.
First, it separates workflow design from workflow execution, a principle
that has proven crucial in reproducible workflow systems such as
\emph{\pkg{batchtools}}
(\citeproc{ref-bischlBatchJobsBatchExperimentsAbstraction2015}{Bischl et
al. 2015}; \citeproc{ref-langBatchtoolsToolsWork2017}{Lang et al. 2017})
and \emph{\pkg{targets}}
(\citeproc{ref-landauTargetsPackageDynamic2021}{Landau 2021}). Second,
it creates a standardized entry point that ensures all engines can be
configured in a consistent manner. The control object thus acts as a
blueprint: it specifies the workflow at a high level, while leaving the
details of computation to the underlying engines. This design choice
reduces complexity by allowing users to adjust workflows by modifying
parameters rather than rewriting code, and it ensures transparency, as
all configuration choices are captured in a single, inspectable object.

The main benefit of the control object paradigm is that it embeds
reproducibility and transparency directly into the workflow definition.
Because configuration parameters are stored in a single object, workflow
specifications can be inspected, shared, and reconstructed more easily.
This principle is consistent with the goals of reproducible workflow
tools in \proglang{R}, such as \emph{\pkg{targets}} which emphasize
explicit pipeline structure and reproducible execution
(\citeproc{ref-landauTargetsPackageDynamic2021}{Landau 2021}). In
addition, the control object enables transparency by making the
structure of the workflow explicit: users can inspect, share, or modify
the object without altering the underlying code base. This separation of
configuration from execution facilitates collaboration, since domain
experts can adjust workflows by editing parameters rather than rewriting
\proglang{R} functions. Beyond this, the control object can be viewed as
a modular construction kit: users may assemble complete workflows from
reusable control components developed by others. This composability
enables collaborative development without rewriting entire control
specifications, strengthening the framework's balance between structure
and openness. At the same time, it promotes flexibility: switching
between different engines---for example, a linear model versus a random
forest for training---requires only a minimal change in the control
object, with no additional code overhead.

Beyond reproducibility and transparency, the control object also
provides a natural entry point for extensibility and automation. Since
engines are selected and parameterized entirely through this object, new
functionality can be added to a workflow by simply registering a new
engine and referencing it in the configuration. No changes to the
workflow execution logic are required, which makes the system resilient
to ongoing methodological innovation. This design mirrors principles of
modular software development, where abstraction boundaries enable
independent evolution of components
(\citeproc{ref-parnasCriteriaBeUsed1972}{Parnas 1972}). In practice, the
control object can also serve as a bridge to automation: because it
captures the full specification of a workflow in a structured format, it
can be versioned, programmatically modified, or integrated into
continuous integration pipelines. This ensures that workflows remain
both adaptable and auditable, supporting sustainable research practices
in fast-moving areas such as fairness, robustness, and explainability.

The control object paradigm also points beyond local workflow
specification toward broader questions of distributed statistical
computing. We use the term distributed in a deliberately broad sense: it
refers not only to computation across distributed resources, but also to
methodological development across distributed contributors and
computational settings. By linking \pkg{flowengineR} to both distributed
computing practice and normative design goals, this architectural
perspective establishes a foundation for extensible and future-proof
methodological work. Distributed computation has long been a central
paradigm in computer science
(\citeproc{ref-gargElementsDistributedComputing2002}{Garg 2002}),
related ideas in statistics can at least be traced back to the DSC
conference series starting in 1999
(\url{https://www.r-project.org/conferences/DSC-1999}). More recently,
the topic has regained impetus with the advent of massively distributed
learning strategies (\citeproc{ref-koInDepthAnalysisDistributed2021}{Ko
et al. 2021}) combined with the need to encapsulate data to be used for
learning at different locations for privacy reasons as in federated
learning (\citeproc{ref-kairouzAdvancesOpenProblems2021}{Kairouz et al.
2021}).

This architectural orientation is deliberate: by structuring workflows
as modular engines with standardized interfaces, \pkg{flowengineR} aims
to support scalable execution and to make computational processes more
transparent and inspectable. These properties are particularly important
in real-world applications, especially in regulated domains, where
workflows must be not only accurate but also reproducible, auditable,
and traceable across computational settings. They also help situate
flowengineR in relation to other workflow infrastructures that address
persistence, coordination, and execution across computational settings.

While the design philosophy of \pkg{flowengineR} shares conceptual roots
with registry-based workflow systems such as \pkg{batchtools}
(\citeproc{ref-langBatchtoolsToolsWork2017}{Lang et al. 2017}), it
pursues a different trade-off between persistence and interactivity.
External registries excel when workflows must persist across sessions or
be dispatched to large computing clusters, as they maintain job states
outside the \proglang{R} process. In contrast, \pkg{flowengineR}
encapsulates all configuration and results within in-memory \proglang{R}
objects---the control object and the result object---to maximize
transparency, portability, and ease of inspection during interactive
analysis. Both paradigms therefore address complementary needs:
persistent, high-throughput execution on the one hand and lightweight,
object-centric reproducibility on the other. Rather than replacing
registry systems, \pkg{flowengineR} offers a compatible and internally
consistent alternative for scenarios where immediate feedback, flexible
configuration, and direct inspectability within \proglang{R} are the
primary goals.


\subsection{\texorpdfstring{Design Principles
}{Design Principles }}\label{sec-2_2}

Building on the desirables introduced in Section~\ref{sec-1}
(\hyperlink{S1}{\Sbadge[S1color]{1}}--\hyperlink{S7}{\Sbadge[S7color]{7}}),
we now derive a set of concrete design principles
(\hyperlink{D1}{\Dbadge[D1color]{1}}--\hyperlink{D12}{\Dbadge[D12color]{12}})
that operationalize these properties within the architecture of
\pkg{flowengineR}. Each principle translates one or more desirables into
actionable guidelines for workflow design, providing the foundation for
a modular, extensible, and reproducible framework.

By abstracting workflow steps into autonomous components,
\pkg{flowengineR} achieves orthogonalization of responsibilities: every
engine can be developed, tested, and maintained independently. This
separation of concerns allows individual modules to evolve without
breaking the surrounding pipeline, a property that is crucial in a
dynamic methodological environment. In addition, outputs generated by
engines are standard \proglang{R} objects that can, in principle, be
stored and reused in later workflows, further strengthening the
flexibility of the approach. This means that a new evaluation engine or
a novel fairness adjustment can be plugged into an existing workflow
without rewriting other parts of the system. The resulting process chain
is transparent, reproducible, and adaptable, creating a foundation for
workflows that remain sustainable in the long run.

\hypertarget{D1}{}

\textbf{\Dbadge[D1color]{1} --- Separation of Concerns ---}
\pkg{flowengineR} decomposes workflows into clearly delimited
responsibilities. Each engine is responsible for a clearly defined task,
such as splitting data, training a model, or evaluating performance, and
interacts with other engines only through standardized input--output
structures. This minimal coupling ensures that engines can be developed,
maintained, or replaced independently, without affecting the surrounding
workflow. The principle reflects established insights from modular
software engineering, where independent components reduce complexity and
foster sustainable growth
(\citeproc{ref-gammaDesignPatternsElements1995}{Gamma 1995};
\citeproc{ref-parnasCriteriaBeUsed1972}{Parnas 1972}). Conceptually,
this corresponds to the \emph{Strategy} and \emph{Template Method}
patterns (\citeproc{ref-gammaDesignPatternsElements1995}{Gamma 1995}),
with workflow steps realized as interchangeable \emph{Strategies} and
the overarching execution logic following a \emph{Template Method}
structure. By enforcing orthogonal process steps
(\hyperlink{S1}{\Sbadge[S1color]{1}}), \pkg{flowengineR} addresses the
lack of architectural openness identified in Section~\ref{sec-1}.
Existing frameworks often embed fairness methods or workflow logic
tightly within fixed pipelines, making extension difficult. In contrast,
\pkg{flowengineR} keeps engines autonomous, enabling flexible
substitution and incremental improvement. This principle forms the
backbone of the framework's modularity and prepares the ground for
extensibility in subsequent design principles.

Beyond this separation of concerns, \textbf{modularity} ensures that
every engine can be replaced independently, supporting long-term
adaptability and lowering maintenance costs. \textbf{Class-free
interfaces} avoid the rigidity of conventional object-oriented
hierarchies, which in \proglang{R} often introduce unnecessary overhead
and reduce flexibility. Instead, engines communicate through simple and
transparent data structures, keeping workflows lightweight while
remaining interoperable. Finally, \textbf{structured input--output with
smart defaults} balances transparency and usability: engines can produce
detailed result objects when required, but defaults are designed to
minimize configuration effort for users. This combination of principles
creates an architecture that is simultaneously open to extension and
efficient in daily use. By lowering technical barriers, it allows both
researchers and practitioners to focus on methodological innovation
rather than infrastructure complexity.

\hypertarget{D2}{}

\textbf{\Dbadge[D2color]{2} --- Transparent Data Flow ---}
\pkg{flowengineR} makes communication between workflow components
explicit from the outset. Communication between engines is explicit,
standardized, and logged, ensuring that every transformation in the
workflow can be traced and reproduced. Instead of opaque or hidden
function calls, data are exchanged as structured list objects that can
be inspected, versioned, and stored. This design echoes practices in
reproducible pipeline frameworks such as \pkg{targets}
(\citeproc{ref-landauTargetsPackageDynamic2021}{Landau 2021}), where
dependency graphs make analytical processes auditable and comparable
across runs. Conceptually, this relates to the \emph{Facade} and
\emph{Adapter} patterns
(\citeproc{ref-gammaDesignPatternsElements1995}{Gamma 1995}), which
provide a uniform interface for interacting components while allowing
heterogeneous or external outputs to be integrated into the standardized
I/O scheme. By emphasizing structured I/O with smart defaults
(\hyperlink{S4}{\Sbadge[S4color]{4}}), \pkg{flowengineR} provides
clarity without imposing unnecessary configuration overhead: users can
rely on defaults for standard workflows, yet retain the option to fully
inspect and override inputs and outputs when needed. In doing so, the
framework directly addresses the research gap identified in Section 1,
where existing infrastructures often lacked transparent communication.
This transparency complements separation of concerns
(\hyperlink{D1}{\Dbadge[D1color]{1}}) by ensuring that modular engines
remain not only independent but also interoperable and accountable.

\hypertarget{D3}{}

\textbf{\Dbadge[D3color]{3} --- Standardized I/O Contract ---} Modular
workflows in \pkg{flowengineR} rely on uniform conventions for how data
and metadata enter and exit each engine. Instead of relying on classes
or bespoke wrappers, every engine communicates through schema-like
structures: inputs and outputs are plain \proglang{R} objects, yet they
follow a consistent, minimal specification. Smart defaults further
reduce boilerplate by automatically populating optional fields, allowing
developers to focus on substantive method design rather than wiring
details. This design guarantees that engines remain interoperable and
composable, since each component can rely on predictable interfaces.
Conceptually, this corresponds to the \emph{Adapter} and \emph{Builder}
patterns (\citeproc{ref-gammaDesignPatternsElements1995}{Gamma 1995}):
the \emph{Adapter} integrates heterogeneous components into a uniform
scheme, while the \emph{Builder} progressively constructs structured
outputs from simple objects without exposing unnecessary complexity. By
aligning with modularity (\hyperlink{S2}{\Sbadge[S2color]{2}}),
class-free interfaces (\hyperlink{S3}{\Sbadge[S3color]{3}}), and
structured I/O (\hyperlink{S4}{\Sbadge[S4color]{4}}), the I/O contract
ensures both flexibility and safety: developers can extend functionality
without breaking pipelines, while users benefit from transparent,
inspectable data flow. Comparable concerns appear in systems such as
\pkg{targets} and \pkg{OpenML}: \pkg{targets} makes pipeline structure
explicit for reproducible execution, while \pkg{OpenML} uses
standardized task descriptions, metadata, APIs, and benchmark suites to
support comparability and reuse
(\citeproc{ref-bischlOpenMLBenchmarkingSuites2021}{Bischl et al. 2021};
\citeproc{ref-landauTargetsPackageDynamic2021}{Landau 2021}).

\hypertarget{D4}{}

\textbf{\Dbadge[D4color]{4} --- OpenEngine Interface ---}
\pkg{flowengineR} connects workflow components through a lightweight,
class-free protocol. Engine roles are fixed at the level of broad
types---such as split, train, evaluate, or report---but remain open for
arbitrary implementations. Within each type, developers can register new
instances that follow a standardized input--output convention, ensuring
immediate compatibility without changes to the surrounding
infrastructure. Conceptually, this corresponds to the \emph{Bridge} and
\emph{Factory} Method patterns
(\citeproc{ref-gammaDesignPatternsElements1995}{Gamma 1995}): the
\emph{Bridge} separates engine roles from their concrete
implementations, while the \emph{Factory} Method governs how new engines
are instantiated and registered. This structure--flexibility balance
allows workflows to be predictable in architecture while remaining open
to methodological innovation. By operationalizing modularity by design
(\hyperlink{S2}{\Sbadge[S2color]{2}}) and class-free interfaces
(\hyperlink{S3}{\Sbadge[S3color]{3}}), this protocol lowers the barrier
for extension: a new training algorithm or evaluation method can be
added as an independent engine instance, plugged directly into existing
pipelines. This design reflects foundational insights on modular
decomposition and extensibility
(\citeproc{ref-parnasCriteriaBeUsed1972}{Parnas 1972}), adapts them to
the functional, class-free idioms of \proglang{R}
(\citeproc{ref-wickhamAdvanced2019}{Wickham 2019}), and builds on
experiences from declarative pipeline frameworks such as \pkg{targets}
(\citeproc{ref-landauTargetsPackageDynamic2021}{Landau 2021}). The
result is an interface that remains lightweight and adaptable, while
still enforcing consistency across workflows.

\hypertarget{D5}{}

\textbf{\Dbadge[D5color]{5} --- FAIR by Design ---} In addition to
openness and modularity, \pkg{flowengineR} is designed to comply with
the FAIR principles of research data management, which demand that data
and analyses be \textbf{Findable, Accessible, Interoperable, and
Reusable} (\citeproc{ref-wilkinsonFAIRGuidingPrinciples2016}{Wilkinson
et al. 2016}). All workflow specifications are stored in structured
control objects, making them easy to share, inspect, and version.
Results are passed between engines through standardized I/O structures,
ensuring interoperability across workflow components. Reproducibility is
supported through explicit workflow specifications, deterministic seeds
where applicable, and structured recording of configuration and outputs.
Conceptually, this resonates with the \emph{Command} and \emph{Memento}
patterns (\citeproc{ref-gammaDesignPatternsElements1995}{Gamma 1995}):
workflows are expressed as explicit commands that can be logged and
replayed, while key states and parameters are preserved as mementos to
support reproducibility and auditing. For distributed execution,
integration with batch-oriented frameworks such as \pkg{batchtools} can
provide additional job-state management and persistence
(\citeproc{ref-langBatchtoolsToolsWork2017}{Lang et al. 2017}). By
embedding reproducibility directly into its architecture,
\pkg{flowengineR} aligns methodological innovation with established
standards for research data management. In doing so, it operationalizes
compatibility with distributed statistical computing
(\hyperlink{S5}{\Sbadge[S5color]{5}}) and regulatory suitability
(\hyperlink{S7}{\Sbadge[S7color]{7}}), ensuring sustainability in both
academic and supervised domains.

\hypertarget{D6}{}

\textbf{\Dbadge[D6color]{6} --- Low-Barrier Extensibility ---} While
modularity and standardized interfaces define the architecture of
\pkg{flowengineR}, its broader impact depends on the \textbf{ease with
which new engines can be added}. The framework deliberately minimizes
boilerplate: a developer need only provide a plain \proglang{R} function
adhering to the standardized I/O contract, and the engine becomes
immediately usable in workflows. To further lower the barrier,
scaffolding functions and documentation templates are provided, offering
ready-made skeletons for common engine types. This approach ensures that
domain experts---such as statisticians or applied researchers---can
contribute novel methods without mastering the internals of the
framework. By linking modularity (\hyperlink{S2}{\Sbadge[S2color]{2}}),
class-free interfaces (\hyperlink{S3}{\Sbadge[S3color]{3}}), and
accessibility (\hyperlink{S6}{\Sbadge[S6color]{6}}), \pkg{flowengineR}
supports a culture of rapid experimentation and open collaboration. This
principle resonates with long-standing ideas on modular decomposition
(\citeproc{ref-parnasCriteriaBeUsed1972}{Parnas 1972}) and design
patterns (\citeproc{ref-gammaDesignPatternsElements1995}{Gamma 1995}),
but adapts them pragmatically to the functional idioms of \proglang{R}
(\citeproc{ref-wickhamAdvanced2019}{Wickham 2019}), yielding a framework
that is both structured and approachable.

\hypertarget{D7}{}

\textbf{\Dbadge[D7color]{7} --- Sanity Checks ---} Beyond modularity and
transparent communication, \pkg{flowengineR} embeds systematic sanity
checks into its workflow execution. These checks verify that workflows
are structurally complete and internally consistent before engines are
executed. For example, they ensure that required splits are present,
that engines receive the expected input structures, and that all
declared outputs can be resolved downstream. Importantly, sanity checks
focus on completeness and coherence, not on the substantive validity of
a statistical method or dataset. By separating structural integrity from
methodological judgment, the framework helps users detect configuration
issues early while keeping workflows lightweight and flexible. This
principle reduces runtime errors, improves the user experience, and
enhances reproducibility by ensuring that pipelines are not only modular
but also systematically verifiable. Conceptually, this is analogous to
the \emph{Decorator} pattern
(\citeproc{ref-gammaDesignPatternsElements1995}{Gamma 1995}), where
additional responsibilities are layered on top of existing components
without altering their core functionality. In operational terms, sanity
checks translate structured I/O with smart defaults
(\hyperlink{S4}{\Sbadge[S4color]{4}}) into safeguards accessible even to
non-expert users (\hyperlink{S6}{\Sbadge[S6color]{6}}), reinforcing
transparency and reliability at the architectural level.

\hypertarget{D8}{}\hypertarget{D9}{}

\textbf{\Dbadge[D8color]{8} --- Testability by Design and
\Dbadge[D9color]{9} --- Continuous Integration \& Cross-Platform
Validation ---} To maintain robustness in an open and extensible
ecosystem, \pkg{flowengineR} embeds systematic \textbf{quality
assurance} at multiple levels. End-to-end tests verify that entire
workflows execute correctly, ensuring that combinations of engines
interact as expected. At the same time, unit tests target individual
engines, providing fine-grained checks for correctness and preventing
regressions when methods evolve. These testing practices are integrated
into \textbf{continuous integration (CI) pipelines}, so that every
change to the framework is automatically validated across environments.
Conceptually, this aligns with the \emph{Command} and \emph{Facade}
patterns (\citeproc{ref-gammaDesignPatternsElements1995}{Gamma 1995}):
test runs can be expressed as \emph{commands} that are logged and
replayed for deterministic validation, while a unified \emph{facade}
provides a consistent interface for invoking heterogeneous test
scenarios. This approach reflects best practices in modular software
engineering and mirrors the increasing use of CI in statistical software
development. For example, Bengtsson
(\citeproc{ref-bengtssonUnifyingFrameworkParallel2021}{2021})
illustrates how a minimal API can support multiple sequential, parallel,
and distributed backends, highlighting the need for systematic backend
validation in extensible execution frameworks. By combining modular
extensibility with strict quality assurance, \pkg{flowengineR} ensures
that new engines introduced by different contributors do not compromise
stability. In operational terms, \hyperlink{D8}{\Dbadge[D8color]{8}} and
\hyperlink{D9}{\Dbadge[D9color]{9}} ensure that modularity by design
(\hyperlink{S2}{\Sbadge[S2color]{2}}) is balanced with reliability,
while regulatory suitability (\hyperlink{S7}{\Sbadge[S7color]{7}}) is
supported through documented and repeatable quality controls.

\hypertarget{D10}{}\hypertarget{D11}{}\hypertarget{D12}{}

\textbf{\Dbadge[D10color]{10} --- Lean Overhead (Efficiency),
\Dbadge[D11color]{11} --- Selective Execution \& Scale (Scalability),
and \Dbadge[D12color]{12} --- Naming \& Traceability Consistency ---}
The final set of design principles addresses the practical requirements
of executing workflows in real-world environments. First,
\pkg{flowengineR} aims to introduce minimal overhead, so that pipelines
remain fast and responsive even when composed of many engines. Second,
workflows can be executed partially and outputs can be restricted to
what is actually needed, avoiding unnecessary computations and storage
costs. This form of selective execution is also emphasized in workflow
systems such as \pkg{targets}, which enables users to skip already
up-to-date steps to reduce runtime overhead
(\citeproc{ref-landauTargetsPackageDynamic2021}{Landau 2021}). By
contrast, node-based workflow environments such as KNIME may rely on
explicit caching or materialization of intermediate tables in certain
workflows, which can be beneficial for interactive inspection but may
increase storage demands depending on configuration and use case
(\citeproc{ref-knimeDocumentation2025}{KNIME AG 2025}). Finally,
systematic labeling and naming conventions provide orientation across
complex workflows. Each engine, split, and output is consistently
tagged, making it straightforward to trace results back to their origin
and to align documentation with execution. Conceptually, these
principles align with the \emph{Iterator}, \emph{Proxy}, and
\emph{Flyweight} patterns
(\citeproc{ref-gammaDesignPatternsElements1995}{Gamma 1995}):
\emph{Iterator} supports traversing workflow components efficiently,
\emph{Proxy} enables transparent remote or distributed execution, and
\emph{Flyweight} minimizes memory use when handling large numbers of
lightweight objects. By operationalizing modularity by design
(\hyperlink{S2}{\Sbadge[S2color]{2}}), distributed statistical computing
compatibility (\hyperlink{S5}{\Sbadge[S5color]{5}}), and regulatory
suitability (\hyperlink{S7}{\Sbadge[S7color]{7}}), these principles
ensure that \pkg{flowengineR} workflows remain not only flexible but
also efficient, scalable, and auditable in practice.

\begin{figure}[H]

\centering{

\includegraphics[width=1\linewidth,height=\textheight,keepaspectratio]{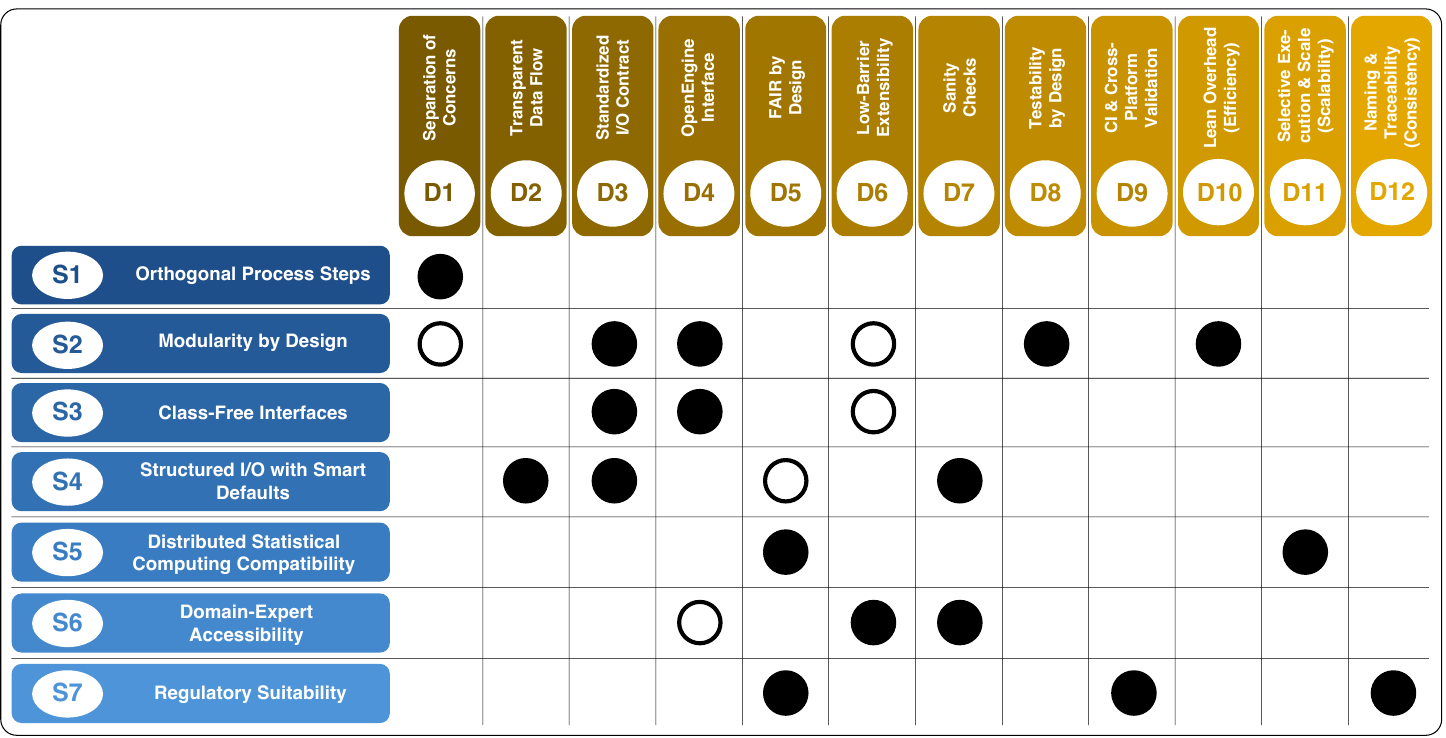}

}

\caption{\label{fig-ds-mapping}\emph{Mapping of design principles
(D1--D12) to desirables (S1--S7). Circles indicate relationships, with
shading encoding strength: primary (filled), supporting (unfilled), and
none.}}

\end{figure}%

Taken together, these principles translate the abstract desirables
introduced in Section~\ref{sec-1} into concrete architectural guidelines
for \pkg{flowengineR}. Their interrelation is summarized in
Figure~\ref{fig-ds-mapping}, which maps each design principle
(\hyperlink{D1}{\Dbadge[D1color]{1}}--\hyperlink{D12}{\Dbadge[D12color]{12}})
to the corresponding desirables
(\hyperlink{S1}{\Sbadge[S1color]{1}}--\hyperlink{S7}{\Sbadge[S7color]{7}})
and thus highlights how the framework systematically operationalizes its
foundational properties.


\subsection{\texorpdfstring{OpenEngine Interface Design
}{OpenEngine Interface Design }}\label{openengine-interface-design}

The \textbf{OpenEngine Interface} provides the central mechanism that
balances structure and flexibility in \pkg{flowengineR}. Engine types
are defined at a fixed set of categories---such as \emph{split},
\emph{train}, \emph{evaluate}, or \emph{report}---which represent the
functional roles within a workflow. Within each category, however, the
interface remains open: users and developers can register new engines
that implement alternative methods without modifying the core system.
This architecture follows principles of plugin-based extensibility
familiar from other software ecosystems, but applies them directly
within the \proglang{R} environment. By separating fixed engine roles
from open implementation slots, \pkg{flowengineR} ensures that workflows
remain both predictable in structure and flexible in content, supporting
sustainable extension over time. This design principle
(\hyperlink{D4}{\Dbadge[D4color]{4}}) directly implements the desirable
property of class-free interfaces (\hyperlink{S3}{\Sbadge[S3color]{3}}),
while reinforcing modularity by design
(\hyperlink{S2}{\Sbadge[S2color]{2}}). By keeping the interface
lightweight and open, \pkg{flowengineR} ensures extensibility without
introducing rigid hierarchies.

\begin{figure}[H]

\centering{

\includegraphics[width=1\linewidth,height=\textheight,keepaspectratio]{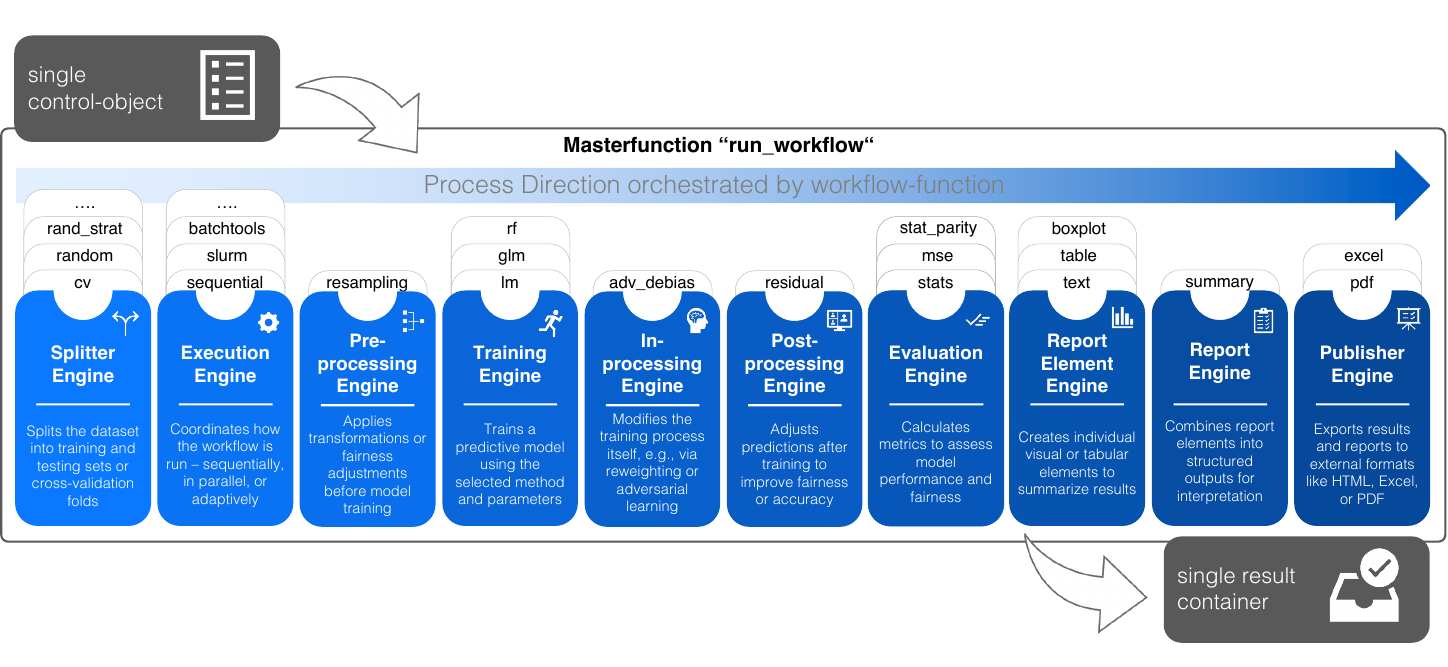}

}

\caption{\label{fig-modular_structure}\emph{Modular Workflow Structure
of flowengineR. The figure illustrates the core engine types (Splitter,
Preprocessing, Inprocessing, Training, Postprocessing, Evaluation,
Reporting, Publishing) and their interaction within a workflow pipeline.
This modular design enables clear separation of concerns and
standardized interfaces across all components.}}

\end{figure}

All engines in \pkg{flowengineR} adhere to a \textbf{standardized
input--output protocol}, which forms the backbone of interoperability.
Regardless of whether an engine performs data splitting, model training,
or evaluation, it receives a structured input object and returns a
standardized output. This uniformity enables engines to be combined
freely without ad hoc adaptations, reducing integration costs and
preventing inconsistencies across workflows. The openness of the
interface can be visualized as a matrix of \emph{engine types} by
\emph{engine instances}: each type (e.g., \emph{train}) represents a
structural slot, while each instance (e.g., \emph{train\_glm},
\emph{train\_rf}) fills the slot with a specific implementation.
Figure~\ref{fig-modular_structure} illustrates this modular workflow
structure, showing how the fixed categories interact and how openness
for extensibility is preserved.

The OpenEngine Interface also lowers the barrier for
\textbf{user-generated engines}, enabling distributed development by
researchers and practitioners. Because the interface enforces consistent
input--output conventions, new engines can be written with minimal
boilerplate code and integrated immediately into existing workflows.
This design principle is intended to support community-driven extension
by allowing contributors to develop engines outside the core framework.
In addition, the interface provides a natural foundation for
\textbf{LLM-assisted code generation}: since the structure of an engine
is simple and standardized, large language models can help generate
prototypes that follow the interface specification. While not dependent
on any specific LLM technology, this illustrates how openness and
simplicity create opportunities for accelerating development through
both human and automated contributions.


\subsection{\texorpdfstring{Default Matching and Argument Resolution
}{Default Matching and Argument Resolution }}\label{default-matching-and-argument-resolution}

\pkg{flowengineR} follows \proglang{R}'s \textbf{native mechanisms for
argument matching}, making the configuration of workflows familiar to
users with standard \proglang{R} experience. Named arguments are
resolved according to \proglang{R}'s established rules, ensuring both
consistency and predictability (e.g., match.call, match.arg in base
\proglang{R}). This choice avoids introducing custom configuration
languages or wrappers, which could increase complexity and reduce
transparency. Instead, engines inherit the clarity of \proglang{R}'s
function interfaces while providing additional safeguards through
standardized validation routines. Users can choose between two
configuration styles: a \textbf{short declarative style}, where
workflows can be specified with minimal arguments and rely on smart
defaults, and a \textbf{fully controlled style}, where all parameters
are explicitly set. This dual approach ensures accessibility for
beginners while offering fine-grained control for advanced users,
striking a balance between ease of use and methodological rigor. By
providing structured I/O with smart defaults, these mechanisms
operationalize the desirable \hyperlink{S4}{\Sbadge[S4color]{4}}. They
are captured in the design principles of transparent data flow
(\hyperlink{D2}{\Dbadge[D2color]{2}}) and sanity checks
(\hyperlink{D7}{\Dbadge[D7color]{7}}), ensuring that workflows remain
both practical and auditable.

A further principle of the framework is \textbf{deferred output
specification}. Output formats are not fixed at the start of the
workflow but are accumulated as \textbf{structured result objects} that
carry rich information across engines. This design enables
extensibility, as downstream engines can access outputs that may not
have been anticipated at the time of workflow creation. For example, an
evaluation engine might reuse model metadata originally produced by a
training engine without additional configuration. The approach is
inspired by \textbf{base \proglang{R} conventions}, where functions such
as \texttt{lm()} return structured list objects containing coefficients,
residuals, and fitted values, rather than a preformatted table
(\citeproc{ref-chambersExtending2016}{Chambers 2016}). In
\pkg{flowengineR}, adopting this principle ensures that results remain
flexible, reusable, and adaptable for future extensions, even in
contexts unforeseen during the original workflow design.


\subsection{\texorpdfstring{Logging and Output Specification
}{Logging and Output Specification }}\label{logging-and-output-specification}

Transparency during execution is supported by an integrated
\textbf{logging system} that writes messages to the console. Each
workflow run generates structured debug paths, allowing users to trace
the sequence of executed engines and inspect intermediate results in a
systematic way. This approach facilitates debugging and provides an
auditable record of the workflow, which is particularly important when
workflows are reused across different environments or in regulated
domains. By adopting best practices from reproducible computing and
leveraging established \proglang{R} logging frameworks (e.g.,
\texttt{futile.logger}), \pkg{flowengineR} ensures that diagnostic
information is available without interfering with the main workflow
outputs.

In addition to global logging, the framework supports \textbf{local
output control} through the \texttt{specific\_output} mechanism. Unlike
global configuration options, this feature is implemented at the
\textbf{engine level}: developers can define a dedicated
\texttt{specific\_output} element within their engine to expose selected
subsets of results. This allows new engines to provide lightweight
outputs for reporting or visualization, while the complete structured
object remains available for reproducibility. Users who require
selective output from existing engines can either extend or duplicate
them to include a \texttt{specific\_output} branch. In this way, local
output control becomes a design decision for engine developers rather
than an external configuration choice. While comparable mechanisms exist
in workflow tools such as \emph{\pkg{targets}}
(\citeproc{ref-landauTargetsPackageDynamic2021}{Landau 2021}) or
\pkg{knitr} (\citeproc{ref-Xie2021-ps}{Xie 2021}), \pkg{flowengineR}
embeds the concept directly into the engine interface, ensuring that
selective output remains consistent with the framework's modular design.
Logging and local output control strengthen transparency and
traceability in line with \hyperlink{D2}{\Dbadge[D2color]{2}}, while
selective execution and controlled outputs support efficiency and
scalability
(\hyperlink{D10}{\Dbadge[D10color]{10}}--\hyperlink{D12}{\Dbadge[D12color]{12}}).
Together, these features enhance regulatory suitability
(\hyperlink{S7}{\Sbadge[S7color]{7}}) by ensuring that workflows remain
reproducible and verifiable.


\subsection{\texorpdfstring{Extensibility by Design
}{Extensibility by Design }}\label{extensibility-by-design}

A defining feature of \pkg{flowengineR} is that \textbf{extensibility is
not an optional add-on but a direct consequence of its architecture}. By
building on principles of modularity, orthogonal process steps, and
standardized interfaces, the framework ensures that new functionality
can be integrated seamlessly without modification of the core. This
distinguishes it from many existing toolkits, where extensibility is
often added as a secondary layer or requires invasive changes. Instead,
\pkg{flowengineR} embeds openness at the architectural level, making
extensibility a structural property rather than a feature. This approach
reflects classic insights from modular software engineering
(\citeproc{ref-parnasCriteriaBeUsed1972}{Parnas 1972}) and adapts them
to plugin-oriented architectures in \proglang{R}.

The openness of the interface allows \textbf{new engines to be developed
and maintained independently of the core system}. Contributors can
implement additional training methods, evaluation metrics, or reporting
formats without altering the central workflow logic, ensuring that
community-driven innovation is immediately usable. This separation of
responsibilities supports sustainable growth: the framework remains
lightweight and stable, while extensions can evolve dynamically under
distributed ownership. In practice, this mirrors the culture of the
\proglang{R} ecosystem, where contributed packages expand functionality
well beyond the original scope of base \proglang{R}. By enabling engines
to function as modular add-ons, \pkg{flowengineR} creates an extensible
ecosystem that can scale through community contributions.

Extensibility in \pkg{flowengineR} also ensures that the framework
remains \textbf{future-proof} as methodological innovations continue to
emerge. Because all engines communicate through open interfaces, novel
methods---such as new fairness metrics, visualization routines, or
publishing formats---can be added seamlessly without disrupting existing
workflows. The framework deliberately separates a \textbf{lightweight,
stable core} from \textbf{flexible extensions}, minimizing dependencies
while maximizing adaptability. This distinction guarantees that the core
provides long-term stability, while extensions supply the evolving
functionality required by diverse research communities. Similar
strategies can be observed in extensible frameworks such as
\emph{\pkg{tidymodels}} and \emph{\pkg{mlr3}}, where modular design
keeps the infrastructure lean while encouraging a rich ecosystem of
add-ons.

\begin{figure}[H]

\centering{

\includegraphics[width=1\linewidth,height=\textheight,keepaspectratio]{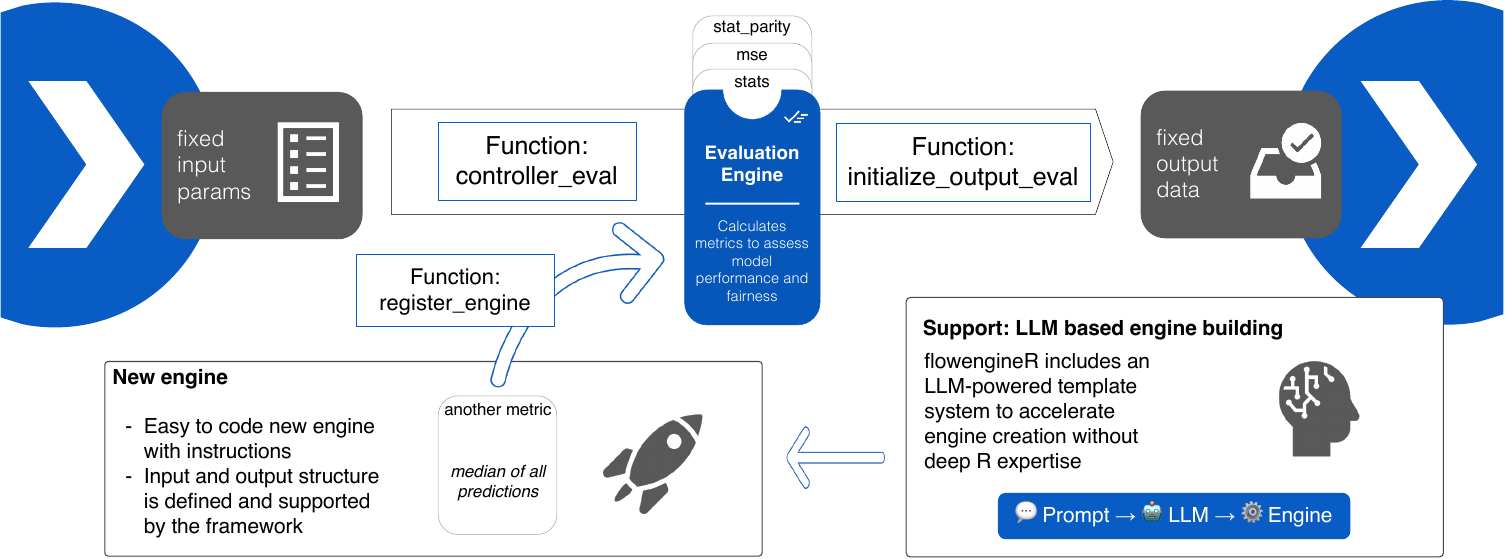}

}

\caption{\label{fig-extensibility}\emph{Extensibility through
standardized engine interfaces. The figure shows how new engines can be
added independently without modifying the core system. This openness
allows community-driven contributions and long-term scalability of the
framework.}}

\end{figure}

The framework is designed to support \textbf{independent evolution}
through user-defined extensions. Because engines are modular and
communicate via standardized interfaces, new instances can be created
without any intervention from the original developers. This distributed
ownership allows the ecosystem to grow autonomously, with specialized
engines emerging from different research communities while remaining
interoperable within the framework. The openness of this design can be
visualized through the \textbf{matrix of engine types and engine
instances}, where each type (e.g., \emph{train}, \emph{evaluate},
\emph{report}) defines a structural slot, and user-defined engines fill
these slots with specific implementations
(Figure~\ref{fig-extensibility}). This perspective highlights
extensibility as a systemic property of the architecture rather than an
afterthought.

\subsection{\texorpdfstring{Relation to Existing Frameworks
}{Relation to Existing Frameworks }}\label{relation-to-existing-frameworks}

\pkg{flowengineR}'s architectural philosophy aligns conceptually with
the modular design of the \pkg{mlr3} ecosystem
(\citeproc{ref-langMlr3ModernObjectoriented2019}{Lang et al. 2019}),
where learners, tasks, and pipeline operators (\pkg{mlr3pipelines},
Binder et al. (\citeproc{ref-mlr3pipelines}{2021})) can be flexibly
combined for advanced modeling and tuning. Likewise, \pkg{mlr3fairness}
(\citeproc{ref-pfistererMlr3fairnessFairnessAuditing2025}{Pfisterer et
al. 2025}) extends this environment with fairness measures and
bias-mitigation operators. In contrast, \pkg{flowengineR} generalizes
modularity \textbf{beyond modeling pipelines}: each stage of an
analytical workflow---data splitting, preprocessing, in-processing,
post-processing, evaluation, and reporting/publishing---is implemented
as a pluggable \textbf{engine} following a unified, class-free I/O
schema. This yields a low-barrier, script-based extensibility model
accessible to domain experts, while \pkg{mlr3} remains a powerful
infrastructure for technically advanced users familiar with R6-based
constructs. The systems are therefore \textbf{complementary rather than
competing}: \pkg{mlr3} provides a rich foundation for modeling and
tuning, whereas \pkg{flowengineR} serves as a lightweight orchestration
layer enabling end-to-end, audit-ready workflows and can embed
\pkg{mlr3} components where appropriate.

Within the broader \proglang{R} ecosystem, \pkg{flowengineR} occupies a
distinct position among workflow tools. Frameworks such as \pkg{targets}
(\citeproc{ref-landauTargetsPackageDynamic2021}{Landau 2021}) and
formerly \pkg{drake} focus on \textbf{reproducible execution graphs} and
task caching, while \pkg{tidymodels}
(\citeproc{ref-kuhnTidymodelsEasilyInstall2018}{Kuhn and Wickham 2018})
emphasizes \textbf{unified model interfaces} and consistent tuning
grammars. \pkg{flowengineR} complements these systems by providing
\textbf{architectural orchestration} rather than task scheduling or
model management: it defines the structure in which individual
tools---potentially including \pkg{targets} pipelines or \pkg{mlr3}
learners---can be embedded as engines within a transparent, auditable
workflow. This orientation bridges the conceptual gap between
execution-level reproducibility and method-level extensibility, offering
a single schema for combining, documenting, and publishing analytical
results.

A broad range of frameworks have been developed to address either
fairness in machine learning or workflow design more generally. These
tools vary widely in scope, extensibility, and the degree to which they
integrate with existing statistical environments. Some emphasize
domain-specific fairness interventions, while others focus on workflow
orchestration or reproducible pipelines. Against this broader
background, flowengineR is best understood as an architectural framework
that aims to connect these concerns within a single workflow logic.
Taken together, these frameworks collectively form a
\textbf{complementary ecosystem} rather than mutually exclusive
alternatives. \pkg{flowengineR} contributes a unifying architectural
layer that connects reproducibility, extensibility, and compliance
within a single workflow logic. Its engine-based structure translates
software-engineering principles such as separation of concerns
(\citeproc{ref-gammaDesignPatternsElements1995}{Gamma 1995}) and the
FAIR paradigm
(\citeproc{ref-wilkinsonFAIRGuidingPrinciples2016}{Wilkinson et al.
2016}) into operational tools for applied research. By lowering
technical barriers for domain experts while maintaining compatibility
with established packages, it extends the reproducible-research
continuum from data processing to publication. Hence, \pkg{flowengineR}
coexists naturally with \pkg{mlr3}, \pkg{targets}, and
\pkg{tidymodels}---serving as an orchestration framework that integrates
and documents their functionality within transparent, end-to-end
analytical workflows.


\section{\texorpdfstring{\textbf{Implementation \& Usage
}}{Implementation \& Usage  }}\label{sec-3}

The latest development version of \pkg{flowengineR} is available on
GitHub:

\begin{Shaded}
\begin{Highlighting}[]
\NormalTok{remotes}\SpecialCharTok{::}\FunctionTok{install\_github}\NormalTok{(}\StringTok{"mwiller1991/flowengineR"}\NormalTok{, }\AttributeTok{build\_vignettes =} \ConstantTok{TRUE}\NormalTok{)}
\FunctionTok{library}\NormalTok{(flowengineR)}
\end{Highlighting}
\end{Shaded}

\noindent A stable release will be submitted to CRAN after the review of
this paper.


\subsection{\texorpdfstring{Layered Engine Architecture
}{Layered Engine Architecture }}\label{layered-engine-architecture}

The architecture of \pkg{flowengineR} is organized into \textbf{five
distinct layers}: \emph{wrappers}, \emph{controllers},
\emph{validators}, \emph{engines}, and \emph{initializers}. Each layer
has a clearly delimited role, ensuring that responsibilities remain
separated and the overall system avoids hidden couplings.
\textbf{Wrappers} provide the high-level interface for users,
translating workflow specifications into standardized calls.
\textbf{Controllers} manage the configuration of engines, supplying
parameter values in a structured format. \textbf{Validators} enforce
input checks and guarantee that the specifications are consistent before
execution. \textbf{Engines} perform the actual computational tasks,
while \textbf{initializers} prepare the objects and structures required
for downstream steps. To support usability, individual engines may
define \textbf{default parameter sets} via dedicated \texttt{default\_*}
functions. These provide sensible values for common use cases while
leaving all parameters open for explicit configuration, ensuring both
accessibility and flexibility. This layered design supports modularity
and standardization, echoing established principles of modular software
engineering, especially the role of well-defined module boundaries
(\citeproc{ref-parnasCriteriaBeUsed1972}{Parnas 1972}), and conceptually
relating to design-pattern ideas such as Facade and Adapter
(\citeproc{ref-gammaDesignPatternsElements1995}{Gamma 1995}). This is
comparable in spirit to modular R modeling ecosystems such as
\emph{\pkg{tidymodels}} and \emph{\pkg{mlr3}}, where functionality is
distributed across interoperating components and packages. In
operational terms, it realizes
\hyperlink{D1}{\Dbadge[D1color]{1}}/\hyperlink{D2}{\Dbadge[D2color]{2}}
(separation of concerns and transparent data flow), thereby translating
\hyperlink{S1}{\Sbadge[S1color]{1}} (Orthogonal Process Steps) and
\hyperlink{S2}{\Sbadge[S2color]{2}} (Modularity by Design) into
practice.

Compatibility across the layers is ensured by \textbf{uniform
input--output (I/O) contracts}, which guarantee that engines can be
combined seamlessly regardless of their internal implementation. Every
wrapper passes standardized input objects to its engine and receives a
structured output in return, which is then validated and initialized for
downstream use. This contract-driven design supports plug-in
compatibility: engines that adhere to the I/O specification can be
integrated into compatible workflows with limited additional adaptation.
The same principle simplifies testing and maintenance, since both
developers and users can rely on predictable data formats. In practice,
this approach echoes ideas from modular programming and workflow systems
such as \emph{\pkg{mlr}}
(\citeproc{ref-bischlMlrMachineLearning2013}{Bischl et al. 2013}) and
\emph{\pkg{targets}}
(\citeproc{ref-landauTargetsPackageDynamic2021}{Landau 2021}), where
reproducibility and composability are ensured through standardized data
structures. By embedding uniform I/O protocols across all layers,
\pkg{flowengineR} makes extensibility a property of the architecture
rather than a matter of convention. This explicitly implements
\hyperlink{D3}{\Dbadge[D3color]{3}} (Standardized I/O Contract) but also
\hyperlink{D4}{\Dbadge[D4color]{4}} (OpenEngine Interface) and
\hyperlink{D8}{\Dbadge[D8color]{8}}/\hyperlink{D9}{\Dbadge[D9color]{9}}
(QA \& CI) while grounding \hyperlink{S3}{\Sbadge[S3color]{3}}
(Class-Free Interfaces) and \hyperlink{S4}{\Sbadge[S4color]{4}}
(Structured I/O with Smart Defaults).


\subsection{\texorpdfstring{Developing Custom Engines
}{Developing Custom Engines }}\label{sec-3_2}

To make the creation of new workflow components accessible,
\pkg{flowengineR} provides \textbf{ready-to-use templates for control
objects}, which serve as the primary entry point for specifying how new
engines interact with the framework. These templates guide developers
through the required configuration fields and illustrate how parameters
are passed, validated, and documented. In addition, the package includes
\textbf{concrete engine files} for existing methods, which can be used
as blueprints for building new engines. A detailed documentation for
each engine type, available in the package vignettes, specifies all
required inputs, outputs, and parameter conventions. Together, these
resources lower the entry barrier for contributors and ensure
consistency across user-developed engines
(\citeproc{ref-wickhamPackages2015}{Wickham 2015}). Aligns with
\hyperlink{D6}{\Dbadge[D6color]{6}} Low-Barrier Extensibility and
therefore contributes to \hyperlink{S6}{\Sbadge[S6color]{6}}
Domain-Expert Accessibility and to \hyperlink{D4}{\Dbadge[D4color]{4}}
OpenEngine Interface, since templates and documentation make new engines
easy to integrate without altering the core.

\begin{Shaded}
\begin{Highlighting}[]
\FunctionTok{register\_engine}\NormalTok{(}\StringTok{"eval\_median"}\NormalTok{, }\StringTok{"\textasciitilde{}/engine\_eval\_median.R"}\NormalTok{)}
\FunctionTok{list\_registered\_engines}\NormalTok{(}\StringTok{"eval"}\NormalTok{)}
\end{Highlighting}
\end{Shaded}

Strong emphasis is placed on \textbf{documentation support}, ensuring
that newly developed engines are transparent, reproducible, and easy to
maintain. Each engine type is accompanied by detailed guidelines and
vignettes that specify parameter requirements, expected inputs and
outputs, and conventions for standardized reporting. Developers are
encouraged to provide rich documentation for each engine, both through
package-level documentation and through vignettes. The latter follows
the tradition of literate programming and dynamic documents in
\proglang{R}, where code, explanation, and output are combined to
support reproducible communication
(\citeproc{ref-xieDynamicDocumentsKnitr2015}{Xie 2015}). By integrating
documentation as a first-class element of engine development,
\pkg{flowengineR} lowers the risk of inconsistencies and provides future
users with clear instructions on how engines are configured, executed,
and extended. This advances \hyperlink{D5}{\Dbadge[D5color]{5}} (FAIR by
Design) and contributes to \hyperlink{S7}{\Sbadge[S7color]{7}}
(Regulatory Suitability) by making specifications fully auditable.

Beyond templates and documentation, \pkg{flowengineR} provides strong
support for \textbf{testing and repository integration}. The package
includes a suite of unit tests that can be adapted for new engines,
allowing developers to validate functionality immediately without having
to set up their own testing infrastructure. Continuous integration
pipelines run these tests automatically, ensuring that contributed
engines remain robust as the framework evolves. In addition, the package
repository contains direct references to template code and testing
scripts, enabling developers to trace examples down to specific files
and lines in the Git version. This combination of automated testing and
transparent repository structure provides a reliable foundation for
community-driven extensions (\citeproc{ref-wickhamPackages2015}{Wickham
2015}). Together, these features anchor
\hyperlink{D8}{\Dbadge[D8color]{8}}/\hyperlink{D9}{\Dbadge[D9color]{9}}
(QA \& CI), bridging \hyperlink{S2}{\Sbadge[S2color]{2}} (modularity)
with \hyperlink{S7}{\Sbadge[S7color]{7}} (auditability).

Finally, engine development in \pkg{flowengineR} follows a principle of
\textbf{closed design}: each engine only accepts the inputs that are
explicitly declared in its specification. This prevents unexpected side
effects and ensures that workflows remain predictable and comparable
across different implementations. By restricting engines to well-defined
interfaces, the framework avoids the ambiguity that often arises when
functions accept arbitrary arguments. This design principle enhances
robustness, simplifies debugging, and guarantees that results can be
reproduced reliably, reflecting established best practices in modular
software engineering (\citeproc{ref-parnasCriteriaBeUsed1972}{Parnas
1972}) and in the design of stable APIs
(\citeproc{ref-wickhamPackages2015}{Wickham 2015}). In terms of
alignment, it sharpens
\hyperlink{D1}{\Dbadge[D1color]{1}}/\hyperlink{D4}{\Dbadge[D4color]{4}}
and strengthens \hyperlink{S7}{\Sbadge[S7color]{7}}, while maintaining
the flexibility of \hyperlink{S2}{\Sbadge[S2color]{2}}.


\subsection{\texorpdfstring{LLM-Assisted Engine Development
}{LLM-Assisted Engine Development }}\label{sec-3_3}

Engines in \pkg{flowengineR} are particularly well-suited for
\textbf{LLM-assisted code generation}, because they share a high degree
of \textbf{structural similarity}. Regardless of whether an engine
performs training, evaluation, or reporting, the surrounding wrapper,
controller, and validator logic follows a consistent pattern. This
repetition makes the underlying design both predictable and easy to
learn, while also reducing the scope of variation that an LLM must
handle. As a result, large language models can be used to draft engine
prototypes from short, well-structured prompts. These drafts are not
treated as validated implementations; they must be reviewed, tested, and
checked against the framework's interface conventions before being
registered or used in workflows. Early explorations in AI-assisted
programming suggest that repetitive code structures, as often found in
\proglang{R} packages, are especially amenable to this form of
automation.

For LLM-assisted development to be effective, certain \textbf{abstract
requirements} must be met. First, interfaces need to be \textbf{clearly
defined}, so that input and output specifications leave no ambiguity for
either humans or automated code generators. Second, engines must adhere
to \textbf{standardized data formats}, which reduces the risk of
inconsistent implementations and makes LLM-assisted prototypes easier to
validate against existing components. Finally, the availability of
\textbf{concise engine templates} provides guidance for both users and
models, reducing the risk of inconsistent implementations. These
requirements are not specific to \pkg{flowengineR} but reflect general
principles of reproducible software design and API clarity. Their
presence within the framework makes engines especially suitable
candidates for automated generation.

The approach is conceptually \textbf{independent of any specific LLM
implementation}. While we illustrate the idea using ChatGPT, the
principle applies equally to other systems capable of code generation.
The benefits are threefold: rapid prototyping of new engines, documented
scaffolding from standardized prompts, and lower entry barriers for
contributors unfamiliar with the full framework internals. This
operationalizes \hyperlink{S6}{\Sbadge[S6color]{6}} Domain-Expert
Accessibility and \hyperlink{D4}{\Dbadge[D4color]{4}} OpenEngine
Interface, since standardized structures make it easy for humans and
LLMs alike to generate new engines. A dedicated helper function
(build\_engine\_from\_llm\_zip()) already implements this concept: it
takes a standardized engine description as input and exports a ZIP
archive containing the necessary instructions and a prompt, allowing
users to easily pass it to any LLM. The accompanying vignette
\emph{``How to Use the LLM Engine Builder''} demonstrates this
end-to-end process using a minimal reproducible example:

\begin{Shaded}
\begin{Highlighting}[]
\FunctionTok{build\_engine\_with\_llm\_zip}\NormalTok{(}\StringTok{"eval"}\NormalTok{,}\StringTok{"The Median of all predictions."}\NormalTok{, }\AttributeTok{zip\_path =} \ConstantTok{NULL}\NormalTok{)}
\CommentTok{\#\textgreater{} LLM zip package created at: /var/folders/m8/fbfgv9pn1t9\_0w\_n38qpk8xh0000gn/T//Rtmp5GFSJZ/llm\_package\_eval.zip}
\CommentTok{\#\textgreater{} }
\CommentTok{\#\textgreater{} To use this ZIP with an LLM (e.g., ChatGPT), follow these instructions:}
\CommentTok{\#\textgreater{} }
\CommentTok{\#\textgreater{} 1. Upload the ZIP file in your chat.}
\CommentTok{\#\textgreater{} 2. Paste the following instruction afterwards:}
\CommentTok{\#\textgreater{} }
\CommentTok{\#\textgreater{} {-}{-}{-} COPY INTO CHAT {-}{-}{-}}
\CommentTok{\#\textgreater{} I have uploaded a ZIP containing a prompt, a working example engine, and a vignette.}
\CommentTok{\#\textgreater{} Please read the prompt first (llm\_prompt\_eval.R). Then carefully review:}
\CommentTok{\#\textgreater{} {-} engine\_eval\_mse.R as a concrete reference implementation}
\CommentTok{\#\textgreater{} {-} detail\_engines\_evaluation.Rmd as documentation of required structure}
\CommentTok{\#\textgreater{} Be precise and complete.}
\CommentTok{\#\textgreater{} Then generate a new engine as specified in the prompt.}
\CommentTok{\#\textgreater{} {-}{-}{-}}
\FunctionTok{register\_engine}\NormalTok{(}\StringTok{"eval\_median"}\NormalTok{, }\StringTok{"\textasciitilde{}/engine\_eval\_median.R"}\NormalTok{)}
\CommentTok{\#\textgreater{} [SUCCESS] Evaluation engine validated successfully.}
\CommentTok{\#\textgreater{} [SUCCESS] Engine registered successfully: eval\_median as type: eval}
\CommentTok{\#\textgreater{} {-}{-}{-}{-}{-}{-}{-}{-}{-}{-}{-}{-}{-}{-}{-}{-}{-}{-}{-}{-}{-}{-}{-}{-}{-}{-}{-}{-}{-}{-}{-}{-}{-}{-}{-}{-}{-}{-}{-}{-}{-}{-}{-}{-}{-}{-}{-}{-}{-}{-}{-}{-}{-}{-}{-}{-}{-}{-}{-}{-}{-}{-}{-}{-}{-}{-}{-}{-}{-}{-}{-}{-}{-}{-}{-}{-}{-}{-}{-}{-}{-}{-}{-}{-}{-}{-}{-}{-}{-}{-}{-}{-}{-}}
\end{Highlighting}
\end{Shaded}

The complete prompt--response transcript, the generated engine
implementation, and the workflow run are archived as an LLM
Demonstration within the flowengineR package (see vignette ``LLM
Demonstration: Eval Median''; all files are accessible under
inst/llm\_demonstrations/2025-08-29\_eval\_median/). This ensures
transparency and reproducibility of the LLM-assisted engine development
process.


\subsection{\texorpdfstring{Engine Registration and Discovery
}{Engine Registration and Discovery }}\label{engine-registration-and-discovery}

The management of engines in \pkg{flowengineR} is handled through a
small set of \textbf{registration functions}. The most central are
\texttt{register\_engine()} and \texttt{list\_registered\_engines()},
which together provide a simple but powerful mechanism for organizing
available components. \texttt{register\_engine()} allows developers or
users to add a new engine by specifying its type (e.g., \emph{train},
\emph{evaluate}) and its location, typically a script path within a
package or user directory. Once registered, the engine becomes part of
the framework's internal registry, ensuring that it can be invoked like
any built-in component. \texttt{list\_registered\_engines()} then
provides a transparent overview of all available engines, supporting
both reproducibility and discoverability by documenting which
implementations are currently available. A minimal example of this
mechanism is shown in \textbf{Code 2} in Section~\ref{sec-3_2}, where a
new evaluation engine is registered and listed.

Engines are not hard-coded into the framework but are instead
\textbf{auto-discovered at runtime}. This means that any engine
registered in a package namespace or in the user's working environment
becomes immediately available without requiring changes to the core
system. The registration process is lightweight and decentralized,
ensuring that new methods can be integrated flexibly by package
developers, research groups, or individual users. This design reflects
concepts of dynamic loading in \proglang{R} and parallels plugin
discovery mechanisms in other ecosystems, such as Python entry points.
By enabling decentralized registration, \pkg{flowengineR} maintains an
open and extensible ecosystem that evolves independently of the central
framework. This embodies \hyperlink{S2}{\Sbadge[S2color]{2}} Modularity
by Design and \hyperlink{D4}{\Dbadge[D4color]{4}} OpenEngine Interface,
since engines can be added without intervention in the core.


\subsection{\texorpdfstring{Publishing and Reporting Extensions
}{Publishing and Reporting Extensions }}\label{publishing-and-reporting-extensions}

Reporting in \pkg{flowengineR} is based on a \textbf{two-layered
architecture} that distinguishes between report elements and reports. At
the first layer, reporting relies on a \textbf{flexible system of
modular report elements}. Each element represents a self-contained
building block --- such as a table, plot, or markdown snippet --- that
transforms standardized workflow outputs into a visual or tabular
representation. Because each element implements the same input--output
schema, users can replace, extend, or combine them freely without
modifying the surrounding workflow. This composability makes it possible
to, for instance, exchange a boxplot for a density plot or add a
fairness summary without regenerating the full report. This flexibility
aligns with the principles of reproducible research found in tools like
knitr (\citeproc{ref-xieDynamicDocumentsKnitr2015}{Xie 2015}) and
rmarkdown (\citeproc{ref-allaireRmarkdownDynamicDocuments2025}{Allaire
et al. 2025}), which integrate computation and documentation in modular
form.

The second layer, the report engines, builds on these elements to
produce complete, publishable artifacts. Reports aggregate multiple
elements, arrange them into layouts, and export them into different
formats such as PDF, HTML, Excel, or JSON. While report elements focus
on content generation, reports manage composition and dissemination.
This architectural separation keeps presentation logic separate from
analytical content, supporting reuse and consistency across workflows.
Similar to \pkg{targets}
(\citeproc{ref-landauTargetsPackageDynamic2021}{Landau 2021}), which
standardizes reproducible execution, \pkg{flowengineR} standardizes
reproducible presentation.

Building upon the report engines, \pkg{flowengineR} also supports
\textbf{publishing extensions} that move beyond visualization or
summarization. These extensions can generate documentation-oriented
reports that support internal review, audit preparation, or
compliance-oriented workflows. By layering publishing engines on top of
reporting engines, the framework allows the same workflow outputs to
serve multiple stakeholders --- from exploratory research to formal
submission. This final layer contributes to
\hyperlink{S7}{\Sbadge[S7color]{7}} (Regulatory Suitability) and
\hyperlink{D12}{\Dbadge[D12color]{12}} (Naming \& Traceability
Consistency) by making results easier to inspect, document, and trace
back to the workflow configuration from which they were generated.


\subsection{\texorpdfstring{Distributed Development and Community
Integration
}{Distributed Development and Community Integration }}\label{distributed-development-and-community-integration}

\pkg{flowengineR} is designed to support \textbf{community-contributed
engines}, lowering the barrier for distributed development and
encouraging broad participation. Because new engines can be implemented
independently and registered without changes to the core, contributors
can extend the framework by adding methods tailored to their own
domains. This openness mirrors the culture of the \proglang{R}
ecosystem, where community packages on CRAN have driven innovation far
beyond the base system. By enabling distributed ownership of extensions,
\pkg{flowengineR} is intended to support an ecosystem in which
methodological advances can be shared and adopted more easily,
reflecting the broader idea of democratized innovation
(\citeproc{ref-hippelDemocratizingInnovation2005}{Hippel 2005}).

The decentralized nature of registration leads to a \textbf{dead drop
registry metaphor}: engines can be shared, published, and discovered
without the need for central ownership or approval. Once an engine is
registered locally, it becomes visible to the framework and is
immediately usable in workflows. This design parallels decentralized
plugin ecosystems in environments such as Python and Java, where modules
can be developed and released independently. The dead drop metaphor
emphasizes autonomy: contributors can publish engines for their
communities, while users can selectively adopt those relevant to their
own workflows. In this way, the registry functions as an open drop-box,
enabling flexible collaboration without bottlenecks in governance.

To sustain long-term growth, \pkg{flowengineR} also adopts an
\textbf{open documentation strategy}. Detailed templates, guidelines,
and vignettes are intended to help external developers onboard more
easily and follow consistent conventions. By making documentation a
central part of the development process, the framework lowers the entry
threshold for new contributors and avoids fragmentation of standards.
This strategy aligns with open science practices and community-driven
initiatives such as ROpenSci, which emphasize transparency and
accessibility in scientific software. As a result, \pkg{flowengineR}
supports distributed contributions while preserving coherence through
shared templates, documentation conventions, and standardized
interfaces. Review, testing, and governance remain necessary to maintain
quality as the ecosystem grows.


\section{\texorpdfstring{\textbf{Use Case: Fairness-Aware Modeling
Workflow }}{Use Case: Fairness-Aware Modeling Workflow }}\label{sec-4}

\subsection{\texorpdfstring{Dataset and Setup
}{Dataset and Setup }}\label{dataset-and-setup}

For this demonstration, we rely on a synthetic banking dataset provided
within the package as test\_data\_2\_base\_credit\_example. The data is
generated through the function \texttt{create\_dataset\_bank()}, which
encodes stylized dependencies between demographic, financial, and
behavioral factors. It includes a binary target variable indicating
whether a credit default occurs, together with socio-demographic
predictors (e.g., gender, age, marital status, region), financial
indicators (e.g., income, loan amount, credit score, loan-to-income
ratio), and derived groupings such as age brackets. The construction of
the dataset follows established principles from credit risk modeling,
where socio-demographic factors, financial attributes, and behavioral
indicators are known to be relevant predictors of default risk
(\citeproc{ref-baesensCreditRiskAnalytics2016}{Baesens et al. 2016};
\citeproc{ref-handStatisticalClassificationMethods1997}{Hand and Henley
1997}). These relationships were encoded into the data-generating
process of \texttt{create\_dataset\_bank()} to create a domain-inspired
synthetic dataset. At the same time, the dataset remains fully
simulated, which avoids confidentiality issues and allows complete
control over the underlying distributions. The use of synthetic data is
increasingly discussed in the machine learning literature as a way to
balance utility, reproducibility, and privacy, while also requiring
careful validation of its limitations
(\citeproc{ref-bellovinPrivacySyntheticDatasets2018}{Bellovin et al.
2018}; \citeproc{ref-breugelSyntheticDataReal2023}{Breugel et al.
2023}). As a benchmark, fairness research has traditionally relied on
widely used datasets such as COMPAS or the Adult census dataset
(\citeproc{ref-friedlerComparativeStudyFairnessenhancing2019}{Friedler
et al. 2019};
\citeproc{ref-hardtEqualityOpportunitySupervised2016a}{Hardt et al.
2016}). By contrast, our synthetic banking dataset provides a
domain-specific alternative for demonstrating credit-risk-oriented
workflow configurations, without claiming to represent real banking
populations.

\begin{figure}[H]

\centering{

\includegraphics[width=1\linewidth,height=\textheight,keepaspectratio]{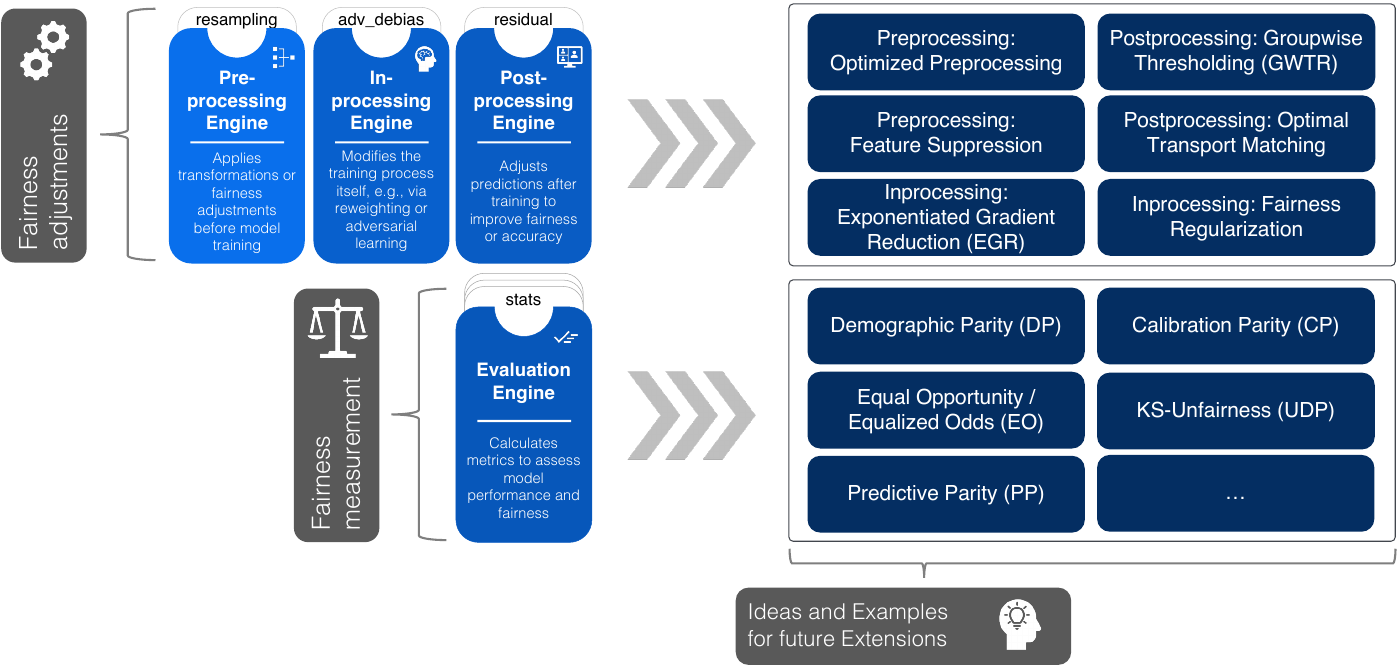}

}

\caption{\label{fig-fairness_usecase}\emph{Fairness as an exemplary use
case in flowengineR. The figure highlights how fairness can be addressed
at different stages of the workflow (Preprocessing, Inprocessing,
Postprocessing) and evaluated using dedicated metrics. This demonstrates
why fairness is an ideal showcase for the framework's modularity.}}

\end{figure}

The scope of this section is deliberately \textbf{focused on workflow
demonstration rather than empirical analysis}. Our objective is not to
extract substantive insights from the data but to show how
\pkg{flowengineR} structures fairness-aware modeling pipelines in a
modular and extensible way. Accordingly, the dataset is kept
intentionally simple, preprocessing steps are minimal, and default
parameterizations are used for most engines. Fairness provides an ideal
showcase for \pkg{flowengineR} because it requires interventions at
different stages of the workflow --- preprocessing, inprocessing,
postprocessing, evaluation, and reporting. This highlights the
modularity of the framework in a way that simpler use cases (e.g., pure
performance optimization) would not. Preprocessing techniques modify the
training data to reduce bias before model fitting, for example by
reweighting or resampling instances. Inprocessing methods directly alter
the learning algorithm, either by adding fairness constraints to the
objective function or by introducing adversarial components to penalize
discriminatory patterns. Postprocessing strategies adjust model outputs
after training, such as by shifting decision thresholds or calibrating
predictions to meet fairness metrics. This tripartite view has become a
standard lens for structuring fairness research, reflecting the
diversity of technical approaches and normative assumptions
(\citeproc{ref-catonFairnessMachineLearning2024}{Caton and Haas 2024}).
Complementary to mitigation, fairness must also be \textbf{measured} in
order to evaluate the success of interventions. Commonly used criteria
include \emph{statistical parity difference}, \emph{equalized odds}, and
\emph{calibration-based measures}
(\citeproc{ref-hardtEqualityOpportunitySupervised2016a}{Hardt et al.
2016}). These metrics provide the benchmarks against which
preprocessing, inprocessing, and postprocessing methods are assessed.
Taken together, the strategies and metrics highlight that fairness
cannot be addressed at a single stage or by a single number alone:
interventions at different workflow levels must be integrated with
systematic evaluation. In the present paper, these categories serve
primarily to motivate the workflow structure used in the use case; the
concrete methodological details of fairness interventions will be
discussed elsewhere. Figure~\ref{fig-fairness_usecase} summarizes this
setup, showing how fairness can be addressed at different points in the
workflow and evaluated consistently across engines.

\subsection{\texorpdfstring{Workflow Configuration
}{Workflow Configuration }}\label{workflow-configuration}

A workflow configuration in \pkg{flowengineR} is handled through a
\textbf{control object}, which specifies the dataset, variables,
engines, and parameters in a structured and reproducible way. The
following \textbf{Code 4} illustrates how a fairness-aware workflow can
be set up on the synthetic credit dataset. It begins with a definition
of the feature, target, and protected variables, followed by the
construction of the control object. This object specifies the training
engine (\texttt{train\_lm}), the postprocessing step
(\texttt{postprocessing\_fairness\_genresidual}), and evaluation engines
(\texttt{eval\_statisticalparity} and \texttt{eval\_mse}). Each section
of the control object is defined transparently, including logging
settings, seeds for reproducibility, and controller functions that merge
user-specified and default parameters. By consolidating these
specifications into a single structured object, the workflow design
remains explicit and auditable. Once defined, the entire workflow can be
executed with a single call to \texttt{run\_workflow(control)}, ensuring
transparency and reproducibility in line with best practices for
reproducible statistical computing.

\begin{Shaded}
\begin{Highlighting}[]
\CommentTok{\#Setting variables fitting to the dataset}
\NormalTok{vars }\OtherTok{=} \FunctionTok{controller\_vars}\NormalTok{(}
  \AttributeTok{feature\_vars =} \FunctionTok{c}\NormalTok{(}\StringTok{"income"}\NormalTok{, }\StringTok{"loan\_amount"}\NormalTok{, }\StringTok{"credit\_score"}\NormalTok{, }
                   \StringTok{"professionEmployee"}\NormalTok{, }\StringTok{"professionSelfemployed"}\NormalTok{, }
                    \StringTok{"professionUnemployed"}\NormalTok{),}
  \AttributeTok{protected\_vars =} \FunctionTok{c}\NormalTok{(}\StringTok{"genderFemale"}\NormalTok{, }\StringTok{"genderMale"}\NormalTok{, }\StringTok{"age"}\NormalTok{),}
  \AttributeTok{target\_var =} \StringTok{"default"}\NormalTok{,       }
  \AttributeTok{protected\_vars\_binary =} \FunctionTok{c}\NormalTok{(}\StringTok{"genderFemale"}\NormalTok{, }\StringTok{"genderMale"}\NormalTok{, }
                            \StringTok{"age\_group.\textless{}30"}\NormalTok{, }\StringTok{"age\_group.30{-}50"}\NormalTok{, }
                            \StringTok{"age\_group.50+"}\NormalTok{)}
\NormalTok{)}

\CommentTok{\#Setting control object}
\NormalTok{control }\OtherTok{\textless{}{-}} \FunctionTok{list}\NormalTok{(}
  \AttributeTok{settings =} \FunctionTok{list}\NormalTok{(    }
    \AttributeTok{log =} \FunctionTok{list}\NormalTok{(}
      \AttributeTok{log\_show =} \ConstantTok{TRUE}\NormalTok{,}
      \AttributeTok{log\_level =} \StringTok{"info"}
\NormalTok{    ),}
    \AttributeTok{global\_seed =} \DecValTok{1}
\NormalTok{  ),}
  \AttributeTok{data =} \FunctionTok{list}\NormalTok{(}
    \AttributeTok{vars =}\NormalTok{ vars,}
    \AttributeTok{full =}\NormalTok{ flowengineR}\SpecialCharTok{::}\NormalTok{test\_data\_2\_base\_credit\_example,}
    \AttributeTok{train =} \ConstantTok{NULL}\NormalTok{,}
    \AttributeTok{test =} \ConstantTok{NULL} 
\NormalTok{  ),}
  \AttributeTok{train =} \StringTok{"train\_lm"}\NormalTok{,}
  \AttributeTok{postprocessing =} \StringTok{"postprocessing\_fairness\_genresidual"}\NormalTok{,}
  \AttributeTok{eval =} \FunctionTok{c}\NormalTok{(}\StringTok{"eval\_statisticalparity"}\NormalTok{, }\StringTok{"eval\_mse"}\NormalTok{),}
  \AttributeTok{params =} \FunctionTok{list}\NormalTok{(}
    \AttributeTok{train =} \FunctionTok{controller\_training}\NormalTok{(}
      \AttributeTok{formula =} \FunctionTok{as.formula}\NormalTok{(}\FunctionTok{paste}\NormalTok{(vars}\SpecialCharTok{$}\NormalTok{target\_var, }\StringTok{"\textasciitilde{}"}\NormalTok{, }
                                 \FunctionTok{paste}\NormalTok{(vars}\SpecialCharTok{$}\NormalTok{feature\_vars, }
                                       \AttributeTok{collapse =} \StringTok{"+"}\NormalTok{), }
                                 \StringTok{"+"}\NormalTok{, }
                                 \FunctionTok{paste}\NormalTok{(vars}\SpecialCharTok{$}\NormalTok{protected\_vars, }
                                       \AttributeTok{collapse =} \StringTok{"+"}\NormalTok{))),}
      \AttributeTok{norm\_data =} \ConstantTok{TRUE}
\NormalTok{    ),}
    \AttributeTok{postprocessing =} \FunctionTok{controller\_postprocessing}\NormalTok{(),}
    \AttributeTok{eval =} \FunctionTok{controller\_evaluation}\NormalTok{()}
\NormalTok{  )}
\NormalTok{)}

\CommentTok{\#running the workflow}
\NormalTok{results }\OtherTok{\textless{}{-}} \FunctionTok{run\_workflow}\NormalTok{(control)}
\end{Highlighting}
\end{Shaded}

A key advantage of this design is that workflows can be \textbf{switched
between different variants without rewriting logic}. Switching between
fairness-aware and fairness-unaware configurations requires only
adjusting engine instances, not rewriting workflow logic --- a central
strength of the modular design. For example, the postprocessing fairness
engine can be removed or replaced with an alternative method, while the
rest of the workflow specification remains unchanged. Similarly,
evaluation engines can be extended to include or exclude fairness
metrics depending on the research question. This approach mirrors
configuration management in frameworks such as \emph{\pkg{mlr3}} and
\emph{\pkg{tidymodels}}, where pipelines can be flexibly reconfigured by
exchanging learners or resampling strategies. In \pkg{flowengineR}, the
ability to toggle between workflow variants in a declarative manner
underscores the extensibility and modularity of the architecture, while
maintaining strict reproducibility of results.


\subsection{\texorpdfstring{Evaluation of Fairness and Performance
}{Evaluation of Fairness and Performance }}\label{evaluation-of-fairness-and-performance}

Evaluation in \pkg{flowengineR} combines \textbf{fairness-oriented
metrics} with traditional measures of predictive accuracy. For fairness,
we consider the \textbf{statistical parity difference (SPD)}. SPD
captures the difference in positive prediction rates between protected
and unprotected groups and is commonly used as a group-fairness measure
related to demographic parity. It provides a simple aggregate indicator
of group-level disparity, although it does not capture all dimensions of
algorithmic fairness (\citeproc{ref-dworkFairnessAwareness2012}{Dwork et
al. 2012};
\citeproc{ref-feldmanCertifyingRemovingDisparate2015a}{Feldman et al.
2015}). For accuracy, we use \textbf{mean squared error (MSE)}, which
quantifies predictive performance and enables direct comparison to
standard modeling benchmarks. Together, these metrics allow us to
evaluate one fairness criterion and one performance criterion within a
single workflow. This dual evaluation illustrates how multiple
evaluation engines can co-exist and provide complementary views on
performance and fairness trade-offs. The current implementation includes
exemplary fairness modules for measurement (statistical parity),
preprocessing (resampling), inprocessing (adversarial debiasing), and
postprocessing (residual adjustment). These methods serve as initial
proof-of-concept implementations to demonstrate how fairness
interventions can be integrated at different workflow stages. A broader
and scientifically validated set of fairness methods will be introduced
in a forthcoming paper dedicated to methodological evaluation within
this framework.

To demonstrate the effect of fairness interventions, we compare
\textbf{fairness-aware workflows} against a \textbf{baseline model
without interventions}. The baseline applies a linear model directly to
the dataset, without postprocessing or fairness adjustments, producing
results that may contain group-level disparities. The fairness-aware
workflow, by contrast, incorporates a residual adjustment in the
postprocessing step, explicitly targeting parity between groups. Results
from these two workflows can be evaluated using the same set of metrics,
providing a transparent and reproducible comparison. This setup reflects
experimental designs common in fairness ML benchmarks, where unfair
baselines are contrasted with mitigation strategies to demonstrate the
effectiveness of different methods. By embedding this comparison into
the declarative workflow structure, \pkg{flowengineR} makes it
straightforward to assess whether fairness interventions achieve their
intended goals.

Finally, the framework supports \textbf{group-specific analysis},
breaking down evaluation results by protected subgroups. For example,
statistical parity can be computed separately for gender and age groups,
highlighting where improvements occur and where trade-offs may persist.
Such subgroup analyses provide a more granular picture of model
behavior, showing whether gains in fairness are evenly distributed or
concentrated in specific groups. This functionality aligns with the
recommendations from recent fairness surveys
(\citeproc{ref-mehrabiSurveyBiasFairness2022}{Mehrabi et al. 2022}),
which emphasize the importance of evaluating interventions across
multiple subpopulations. By integrating subgroup-specific reporting into
the workflow, \pkg{flowengineR} ensures that fairness is not only
considered in aggregate but also inspected at the level where
disparities often manifest most strongly.


\subsection{\texorpdfstring{Reporting and Visualization
}{Reporting and Visualization }}\label{reporting-and-visualization}

The reporting system in \pkg{flowengineR} allows workflows to generate a
variety of \textbf{modular outputs}, ranging from tables and plots to
markdown text elements. These reporting engines can be combined
flexibly, enabling users to tailor reporting pipelines to the needs of
different audiences. For example, one engine may generate a table
summarizing fairness and accuracy metrics, while another produces
visualizations such as boxplots of predictions across groups or trend
plots of error rates. Fairness is particularly suitable for
demonstrating the reporting system, since visualizations (e.g.,
group-level boxplots) immediately convey trade-offs that are otherwise
hidden in scalar performance scores. By supporting different output
modalities, \pkg{flowengineR} provides a consistent structure for both
exploratory summaries and more formal reporting.

A central principle of the framework is \textbf{reproducible reporting},
implemented through markdown-based document generation. Each report
element can be embedded in an \proglang{R} Markdown or Quarto pipeline,
so that textual explanations, tables, and figures can be generated
directly from the underlying workflow outputs. This approach aligns with
established best practices in reproducible research
(\citeproc{ref-xieDynamicDocumentsKnitr2015}{Xie 2015}), where analysis
and documentation are tightly integrated. By combining modular reporting
engines with markdown rendering, \pkg{flowengineR} supports reproducible
and transparent communication of results. Users can generate lightweight
markdown summaries for internal review or extend the system with
publishing engines that produce documentation-oriented reports in PDF,
HTML, or other formats. In this way, reporting and visualization become
an integral part of the workflow, bridging the gap between
methodological development and transparent communication of results.


\subsection{\texorpdfstring{Runtime and Efficiency
}{Runtime and Efficiency }}\label{runtime-and-efficiency}

A common concern with modular frameworks is whether additional
abstraction layers introduce computational overhead. In the benchmark
configurations considered here, runtime comparisons indicate that this
overhead is small relative to the runtime of model training and
processing steps. Each engine executes its core task directly, with
wrappers, controllers, and validators adding only lightweight checks and
standardized data handling. Engine-wise runtime analysis suggests that
training and processing steps consume the majority of time in the
evaluated settings, while framework-level coordination contributes only
a limited share. This is consistent with the expectation that
modularity, when carefully designed, does not slow down computation but
instead makes workflows more transparent and auditable. Thus, in the
evaluated benchmark settings, the separation into engines and supporting
layers provides clarity and flexibility without dominating computational
cost.

\begin{figure}[H]

\centering{

\includegraphics[width=1\linewidth,height=\textheight,keepaspectratio]{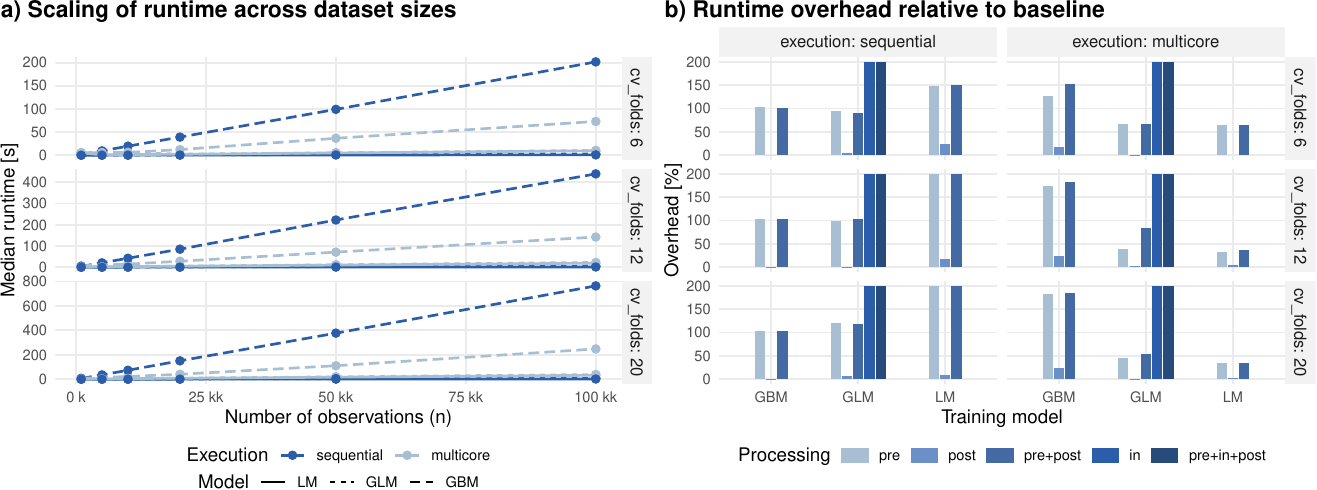}

}

\caption{\label{fig-runtime-scaling}Runtime scaling (a) and additional
computational cost of fairness processing (b). Both panels summarize
median execution times by model type, execution mode, and number of CV
folds. Overhead values are capped at 200 \% for readability.}

\end{figure}%

\vspace{-2em}

\begin{table}[H]

\caption{\label{tbl-runtime-summary}Runtime per 1 000 rows by model and
processing method. The table compares sequential and multicore execution
modes across dataset sizes (XS = 1 000, S = 5 000, M = 10 000, L = 20
000, XL = 50 000, XXL = 100 000) and shows median execution times in
seconds per 1 000 observations; lower values indicate higher
efficiency.}

\centering{

\fontsize{5.0pt}{6.0pt}\selectfont
\begin{tabular*}{\linewidth}{@{\extracolsep{\fill}}l|rrrrrrrrrrrr}
\toprule
 & \multicolumn{6}{c}{sequential} & \multicolumn{6}{c}{multicore} \\ 
\cmidrule(lr){2-7} \cmidrule(lr){8-13}
 & XS & S & M & L & XL & XXL & XS & S & M & L & XL & XXL \\ 
\midrule\addlinespace[2.5pt]
\multicolumn{13}{l}{GBM} \\[2.5pt] 
\midrule\addlinespace[2.5pt]
none & 4.00 & 4.14 & 4.24 & 4.25 & 4.45 & 4.40 & 6.84 & 2.08 & 1.46 & 1.41 & 1.42 & 1.42 \\ 
pre & 8.03 & 8.49 & 8.80 & 8.80 & 9.03 & 8.94 & 7.21 & 2.71 & 2.58 & 2.80 & 3.05 & 3.90 \\ 
post & 4.05 & 4.14 & 4.25 & 4.25 & 4.44 & 4.40 & 6.84 & 2.09 & 1.46 & 1.47 & 1.61 & 1.76 \\ 
pre+post & 8.07 & 8.48 & 8.72 & 8.78 & 9.04 & 8.94 & 7.21 & 2.71 & 2.60 & 2.90 & 3.38 & 4.03 \\ 
\midrule\addlinespace[2.5pt]
\multicolumn{13}{l}{GLM} \\[2.5pt] 
\midrule\addlinespace[2.5pt]
none & 0.09 & 0.08 & 0.07 & 0.06 & 0.06 & 0.06 & 0.49 & 0.45 & 0.33 & 0.26 & 0.23 & 0.23 \\ 
pre & 0.22 & 0.13 & 0.12 & 0.12 & 0.11 & 0.11 & 5.53 & 1.51 & 0.92 & 0.35 & 0.31 & 0.31 \\ 
in & 1.56 & 1.26 & 1.22 & 1.23 & 1.29 & 1.37 & 6.69 & 1.65 & 1.10 & 0.79 & 0.73 & 0.78 \\ 
post & 0.09 & 0.06 & 0.06 & 0.06 & 0.06 & 0.06 & 5.48 & 0.43 & 0.32 & 0.26 & 0.24 & 0.23 \\ 
pre+post & 0.14 & 0.13 & 0.12 & 0.12 & 0.11 & 0.11 & 5.51 & 1.59 & 0.92 & 0.35 & 0.46 & 0.42 \\ 
pre+in+post & 2.56 & 2.31 & 2.26 & 2.32 & 2.60 & 2.75 & 6.91 & 2.20 & 1.48 & 1.22 & 1.59 & 1.89 \\ 
\midrule\addlinespace[2.5pt]
\multicolumn{13}{l}{LM} \\[2.5pt] 
\midrule\addlinespace[2.5pt]
none & 0.04 & 0.04 & 0.02 & 0.02 & 0.01 & 0.01 & 0.44 & 0.42 & 0.31 & 0.24 & 0.20 & 0.19 \\ 
pre & 0.07 & 0.04 & 0.05 & 0.05 & 0.04 & 0.04 & 5.49 & 0.45 & 0.33 & 0.26 & 0.27 & 0.25 \\ 
post & 0.04 & 0.02 & 0.02 & 0.02 & 0.01 & 0.01 & 5.41 & 0.23 & 0.31 & 0.24 & 0.20 & 0.20 \\ 
pre+post & 0.07 & 0.04 & 0.05 & 0.05 & 0.04 & 0.04 & 5.49 & 0.46 & 0.33 & 0.27 & 0.27 & 0.26 \\ 
\bottomrule
\end{tabular*}
\begin{minipage}{\linewidth}
\emph{Values in seconds per 1 000 observations (lower is faster).}\\
\end{minipage}

}

\end{table}%

Figure~\ref{fig-runtime-scaling} and Table~\ref{tbl-runtime-summary}
summarize runtime behavior for the benchmark settings considered in this
paper and suggest that framework-level overhead remains limited in these
configurations. Runtime scales approximately linearly with dataset size,
and multicore execution substantially reduces wall-clock time without
affecting reproducibility. Even for fairness-aware configurations (pre-,
in-, or postprocessing), the relative overhead remains moderate,
suggesting that modularization does not dominate runtime in the
evaluated settings. Interestingly, in multicore mode the relative
runtime per 1 000 observations decreases with larger datasets because
the fixed coordination overhead of \pkg{batchtools} is amortized across
more data, further improving parallel efficiency
(\citeproc{ref-bengtssonUnifyingFrameworkParallel2021}{Bengtsson 2021};
\citeproc{ref-bischlBatchJobsBatchExperimentsAbstraction2015}{Bischl et
al. 2015}; \citeproc{ref-langBatchtoolsToolsWork2017}{Lang et al.
2017}). These results empirically support Design Principles
\hyperlink{D10}{\Dbadge[D10color]{10}} \textbf{Lean Overhead
(Efficiency)} and \hyperlink{D11}{\Dbadge[D11color]{11}}
\textbf{Selective Execution \& Scale (Scalability)}. The complete
runtime test is archived within the flowengineR package (all files are
accessible under inst/runtime\_benchmarks/2025-09-09\_bank\_runtime/).

Efficiency also extends to \textbf{memory management and data handling}.
The framework ensures that only the information necessary for downstream
steps is carried forward, avoiding redundant storage of intermediate
results. For example, while native \texttt{caret::train()} returns a
complex object with many nested elements, \texttt{engine\_train\_*}
functions in \pkg{flowengineR} store only the essential model outputs
and metadata required for later stages. This design reduces memory usage
and keeps workflow objects lightweight, particularly when multiple
models or resampling procedures are run in parallel. Within the
evaluated settings, these results are consistent with the design
principle of efficiency established in Section~\ref{sec-2_2}. Moreover,
the streamlined handling of data and outputs allows \pkg{flowengineR} to
scale gracefully, ensuring that flexibility and extensibility are
achieved without structural slowdown.


\section{\texorpdfstring{\textbf{Generalization to Other Use Cases and
Future Engine Ideas
}}{Generalization to Other Use Cases and Future Engine Ideas }}\label{sec-5}

\subsection{\texorpdfstring{Explainability
}{Explainability }}\label{explainability}

Explainability addresses the need to make model predictions transparent
through \textbf{feature attributions} (e.g., SHAP values, Lundberg and
Lee (\citeproc{ref-lundbergUnifiedApproachInterpreting2017}{2017})),
partial dependence, or residual diagnostics. In regulated domains, such
interpretability is crucial for trust and accountability. Within
\pkg{flowengineR}, explainability can be implemented as
\textbf{evaluation engines} that calculate attribution scores and
diagnostics, and as \textbf{reporting engines} that visualize them in
reproducible tables or plots. Typical metrics include mean absolute SHAP
values or stability of attribution rankings across resamples. This
design follows \hyperlink{S6}{\Sbadge[S6color]{6}} \textbf{Domain-Expert
Accessibility}, since explainability often targets non-technical
stakeholders, and \hyperlink{D2}{\Dbadge[D2color]{2}}
\textbf{Transparent Data Flow}, supporting explicit logging and
inspection of attribution results. The modular engine structure allows
interpretability methods to be plugged directly into workflows without
specialized pipelines. As emphasized by Molnar
(\citeproc{ref-molnarInterpretableMachineLearning2022}{2022}),
explainability methods are an important part of model analysis and
communication; \pkg{flowengineR} provides a structure in which such
methods can be integrated as workflow components.


\subsection{\texorpdfstring{Robustness \& Stress Testing
}{Robustness \& Stress Testing }}\label{robustness-stress-testing}

Robustness and stress testing focus on whether models remain stable
under \textbf{perturbations of inputs} or shifts in data. In
\pkg{flowengineR}, this can be implemented through \textbf{preprocessing
engines} that inject noise, resample data, or create adversarial
variants, and through \textbf{adaptive execution engines} that
repeatedly rerun workflows to monitor variation in key metrics. Common
indicators include the variance of error measures across resamples or
the sensitivity of predictions to structured perturbations. This
supports \hyperlink{S5}{\Sbadge[S5color]{5}} \textbf{Distributed
Statistical Computing Compatibility}, as robustness checks often require
repeated runs, and aligns with \hyperlink{D7}{\Dbadge[D7color]{7}}
\textbf{Sanity Checks}, by helping workflows remain structurally
consistent while stress tests are applied. By embedding robustness
procedures into modular engines, \pkg{flowengineR} allows stability
assessments to be incorporated into workflow evaluation rather than
implemented as separate ad hoc scripts. This is related to, but broader
than, adversarial robustness research
(\citeproc{ref-goodfellowExplainingHarnessingAdversarial2015}{Goodfellow
et al. 2015}).


\subsection{\texorpdfstring{Compliance and Auditing
}{Compliance and Auditing }}\label{sec-5.3}

Compliance-oriented workflows require transparent documentation of data,
models, configuration choices, and results. In \pkg{flowengineR},
\textbf{publishing engines} can support this goal by generating
structured documentation reports that include seeds, parameters, and
results in formats such as PDF or Excel. Complementary
\textbf{evaluation engines} can check adherence to policy-based rules,
e.g., predefined constraints in banking or healthcare. Relevant
indicators include completeness of configuration logs or consistency of
evaluation outcomes across runs. This supports
\hyperlink{S7}{\Sbadge[S7color]{7}} \textbf{Regulatory Suitability} by
making workflows easier to document, inspect, and review. It also builds
on \hyperlink{D5}{\Dbadge[D5color]{5}} \textbf{FAIR by Design},
supporting outputs that are findable, accessible, interoperable, and
reusable. By integrating documentation and rule-based checks into the
workflow, \pkg{flowengineR} can support methodological rigor and
compliance-oriented review.

The importance of such documentation and review capabilities is also
reflected in established risk-management and compliance frameworks in
regulated sectors. In banking, for example, BCBS 239 requires accurate,
complete, timely, and adaptable risk data aggregation and reporting
capabilities, including under stress or crisis conditions
(\citeproc{ref-bcbs2392013}{\emph{Principles for effective risk data
aggregation and risk reporting} 2013}). Similarly, the ECB's RDARR
guidance emphasizes governance, data quality, documentation, reporting
capabilities, and supporting IT infrastructure as prerequisites for
effective supervisory reporting (\citeproc{ref-ecbRDARR2024}{\emph{Guide
on effective risk data aggregation and risk reporting} 2024}). National
regulations such as Germany's BAIT extend these expectations to IT
infrastructures by requiring traceability and documentation of data and
models (\citeproc{ref-bait2021}{BaFin 2021}). While such frameworks do
not directly prescribe fairness metrics, they illustrate why
reproducible, auditable, and well-documented workflow infrastructures
are essential in sensitive application contexts.


\subsection{\texorpdfstring{Future Engine Ideas
}{Future Engine Ideas }}\label{future-engine-ideas}

Finally, the framework is designed to accommodate \textbf{future
extensions}, demonstrating its flexibility beyond the current set of
engines. Possible developments include training engines that integrate
with \textbf{TensorFlow} via the \proglang{R} interface
(\citeproc{ref-allaireTensorflowInterfaceTensorFlow2025}{Allaire and
Tang 2025}), allowing deep learning methods to be embedded into the
workflow while maintaining standardized outputs. On the execution side,
engines compatible with \textbf{SLURM or GPU clusters} could extend
scalability to high-performance computing environments, building on
ideas already present in \emph{\pkg{batchtools}}. These examples
illustrate that the framework is not restricted to its initial engines
but can be extended toward more advanced computational backends,
reflecting the intended extensibility of the architecture.


\section{\texorpdfstring{\textbf{Conclusion and Outlook
}}{Conclusion and Outlook }}\label{sec-6}

\subsection{\texorpdfstring{Summary of Contributions
}{Summary of Contributions }}\label{summary-of-contributions}

The central contribution of \pkg{flowengineR} is the provision of a
framework for \textbf{modular, reproducible, and fairness-aware machine
learning workflows}. By decomposing workflows into standardized engines,
the framework helps keep methodological steps---data splitting,
preprocessing, training, postprocessing, evaluation, and
reporting---transparent, traceable, and replaceable. This modularity
allows researchers to experiment with alternative methods without
rewriting workflow logic, addressing the research gap identified in
Section~\ref{sec-1} regarding architectural openness. At the same time,
\pkg{flowengineR} integrates fairness-aware methods as first-class
citizens, demonstrating how bias mitigation and fairness evaluation can
be embedded across different stages of the pipeline. In contrast to
existing frameworks such as \emph{AIF360}
(\citeproc{ref-bellamyAIFairness3602019}{Bellamy et al. 2019}) or
\emph{Fairlearn} (\citeproc{ref-Weertsfairlearn2023}{Weerts et al.
2023}), which primarily focus on fairness algorithms, \pkg{flowengineR}
emphasizes \textbf{infrastructure and extensibility}, positioning itself
as a meta-framework for integrating methodological innovation within a
reproducible workflow system.

A key strength of the framework lies in its \textbf{structured yet
extensible system design}. The architecture is grounded in a layered
engine system and a declarative control object, which together enforce
standardization while leaving space for extensions. Engine types are
fixed, providing stability, but engine instances are open, enabling new
methods to be registered and integrated seamlessly. Conceptually, this
architecture resonates with established design patterns such as
\emph{Strategy}, \emph{Template Method}, and \emph{Bridge}
(\citeproc{ref-gammaDesignPatternsElements1995}{Gamma 1995}), which
together capture how workflow steps are interchangeable, orchestrated by
a high-level template, and decoupled from their implementations. This
combination of structure and openness supports reproducibility while
enabling community-driven extension. The extensibility is not an
afterthought but a direct consequence of the design principles described
in Section~\ref{sec-2}, allowing new engines to be contributed
independently, provided that they follow the framework's interface
conventions. In this way, \pkg{flowengineR} functions as a lightweight
but expandable backbone, comparable in spirit to modular workflow
frameworks in \proglang{R} such as \pkg{mlr3}
(\citeproc{ref-langMlr3ModernObjectoriented2019}{Lang et al. 2019}) or
\pkg{targets} (\citeproc{ref-landauTargetsPackageDynamic2021}{Landau
2021}), but distinguished by its explicit integration of fairness and
reporting engines.

A further structural contribution of this paper is the explicit,
principle-guided development of the framework. Rather than presenting
\pkg{flowengineR} merely as a collection of functions, the article
identifies the architectural requirements of sustainable workflow
systems and translates them into named design principles. Each major
design choice is therefore made explicit, linked to a corresponding
principle, and connected back to the overall requirements of modularity,
reproducibility, extensibility, accessibility, and compliance. This
makes the framework not only technically extensible but also
conceptually inspectable: readers can trace how abstract design goals
are operationalized in concrete implementation decisions.

Finally, \pkg{flowengineR} is designed to be \textbf{accessible for
domain experts}, not just computer scientists or software engineers. By
leveraging familiar \proglang{R} idioms, native argument matching, and
smart defaults, the framework is intended to lower the entry barrier for
users who may lack deep programming expertise. Templates, documentation,
and reproducible reporting pipelines further support this accessibility
by helping statisticians, social scientists, or practitioners in applied
domains implement methodological extensions. This accessibility is
complemented by transparency: workflow steps are logged and structured
in ways that support reproducibility, inspection, and communication
beyond the original development team.

To conclude this section, \pkg{flowengineR} \textbf{operationalizes the
seven proposed desirables
(\hyperlink{S1}{\Sbadge[S1color]{1}}--\hyperlink{S7}{\Sbadge[S7color]{7}})}
introduced in Section~\ref{sec-1_4} as the architectural requirements of
sustainable workflow frameworks. Figure~\ref{fig-ds-mapping} illustrates
the systematic mapping between these desirables and the twelve design
principles
(\hyperlink{D1}{\Dbadge[D1color]{1}}--\hyperlink{D12}{\Dbadge[D12color]{12}})
that underpin the framework. Each desirable is realized through concrete
design decisions:

\par\vspace{0.8em}

\begin{itemize}
\tightlist
\item
  \hyperlink{S1}{\Sbadge[S1color]{1}} \textbf{-- Orthogonal Process
  Steps:} Implemented through \hyperlink{D1}{\Dbadge[D1color]{1}}
  (\emph{Separation of Concerns}), ensuring each engine executes an
  isolated task and communicates only through standardized interfaces.
\item
  \hyperlink{S2}{\Sbadge[S2color]{2}} \textbf{-- Modularity by Design:}
  Achieved by \hyperlink{D1}{\Dbadge[D1color]{1}},
  \hyperlink{D3}{\Dbadge[D3color]{3}},
  \hyperlink{D4}{\Dbadge[D4color]{4}},
  \hyperlink{D6}{\Dbadge[D6color]{6}},
  \hyperlink{D8}{\Dbadge[D8color]{8}} and
  \hyperlink{D10}{\Dbadge[D10color]{10}}, allowing engines to be
  replaced, extended, or combined without modifying the core system.
\item
  \hyperlink{S3}{\Sbadge[S3color]{3}} \textbf{-- Class-Free Interfaces:}
  Addressed by \hyperlink{D3}{\Dbadge[D3color]{3}} (\emph{Standardized
  I/O Contract}), \hyperlink{D4}{\Dbadge[D4color]{4}} (\emph{OpenEngine
  Interface}) and \hyperlink{D6}{\Dbadge[D6color]{6}} (\emph{Low-Barrier
  Extensibility}), replacing rigid class hierarchies with functional
  transparency.
\item
  \hyperlink{S4}{\Sbadge[S4color]{4}} \textbf{-- Structured
  Input--Output with Smart Defaults:} Realized through
  \hyperlink{D2}{\Dbadge[D2color]{2}},
  \hyperlink{D3}{\Dbadge[D3color]{3}},
  \hyperlink{D5}{\Dbadge[D5color]{5}} and
  \hyperlink{D7}{\Dbadge[D7color]{7}}, balancing transparency, ease of
  use, and systematic validation.
\item
  \hyperlink{S5}{\Sbadge[S5color]{5}} \textbf{-- Distributed Statistical
  Computing Compatibility:} Supported by
  \hyperlink{D5}{\Dbadge[D5color]{5}} (\emph{FAIR by Design}) and
  \hyperlink{D11}{\Dbadge[D11color]{11}} (\emph{Selective Execution \&
  Scale}), enabling reproducible workflows across sequential and
  parallel environments.
\item
  \hyperlink{S6}{\Sbadge[S6color]{6}} \textbf{-- Domain-Expert
  Accessibility:} Fulfilled by \hyperlink{D4}{\Dbadge[D4color]{4}},
  \hyperlink{D6}{\Dbadge[D6color]{6}} (\emph{Low-Barrier Extensibility})
  and \hyperlink{D7}{\Dbadge[D7color]{7}} (\emph{Sanity Checks}), which
  lower technical entry barriers while maintaining quality assurance.
\item
  \hyperlink{S7}{\Sbadge[S7color]{7}} \textbf{-- Regulatory
  Suitability:} Supported through \hyperlink{D5}{\Dbadge[D5color]{5}}
  (\emph{FAIR by Design}), \hyperlink{D9}{\Dbadge[D9color]{9}}
  (\emph{Cross-Platform Validation}), and
  \hyperlink{D12}{\Dbadge[D12color]{12}} (\emph{Naming \& Traceability
  Consistency}), which make workflows easier to document, inspect, and
  review in compliance-oriented settings.
\end{itemize}

\par\vspace{0.8em}

Together, these relationships show how \pkg{flowengineR}
\textbf{addresses the seven desirables} through concrete design
principles, thereby closing the conceptual loop introduced in
Section~\ref{sec-1_4}. By grounding every architectural property in
corresponding design principles, the framework provides a consistent and
verifiable link between abstract requirements and concrete
implementation.


\subsection{\texorpdfstring{Limitations and Future Work
}{Limitations and Future Work }}\label{limitations-and-future-work}

As with any framework, \pkg{flowengineR} comes with \textbf{current
limitations} that must be acknowledged. A central technical challenge
concerns \textbf{versioning of interfaces}. While engine types are
fixed, the possibility of evolving input--output structures raises the
question of how to maintain backward compatibility when interfaces
change. Without clear versioning strategies, users may face unexpected
incompatibilities between engines developed at different points in time.
Similarly, although standardized input--output protocols reduce the risk
of integration failures, edge cases can still arise when multiple
independently developed engines are combined. These issues are well
known in software engineering, where API stability and backward
compatibility are constant concerns. Addressing them in the context of a
distributed and community-driven ecosystem will be an ongoing task,
requiring both governance mechanisms and technical safeguards to ensure
robustness as the framework grows. Beyond technical integration, future
work will focus on systematic validation and benchmarking of fairness
interventions implemented as new engines within the framework. This
includes standardized reference workflows for comparing preprocessing,
inprocessing, and postprocessing methods under identical conditions.
Such benchmark designs would support empirical assessment and
reproducibility of methodological extensions across research groups,
thereby strengthening the scientific reliability of fairness evaluations
conducted with \pkg{flowengineR}.

Looking ahead, a major opportunity for development lies in the
\textbf{integration with other machine learning ecosystems in R}. While
the current focus is on engines implemented specifically for
\pkg{flowengineR}, future work should explore bridges to frameworks such
as \emph{\pkg{tidymodels}}
(\citeproc{ref-kuhnTidymodelsEasilyInstall2018}{Kuhn and Wickham 2018})
and \emph{\pkg{mlr3}}
(\citeproc{ref-langMlr3ModernObjectoriented2019}{Lang et al. 2019}).
Both ecosystems already provide a wealth of learners, resampling
strategies, and evaluation metrics that could enrich the functionality
of \pkg{flowengineR}. Achieving interoperability would allow users to
leverage existing components while benefiting from the modular workflow
architecture and fairness extensions introduced here. More generally,
aligning with the wider \proglang{R} ecosystem will help ensure
long-term sustainability and adoption, avoiding the risk of isolation
and duplication of effort. Interoperability is therefore not only a
technical goal but also a strategic priority for the framework.

Finally, \textbf{emerging opportunities} highlight how \pkg{flowengineR}
can evolve alongside broader trends in statistical computing. In
particular, large language models (LLMs) create possibilities for
semi-automated development of new engines. As demonstrated in
Section~\ref{sec-3_3}, the structural similarity of engines makes them
especially suitable for LLM-assisted scaffolding. Future work may
explore dedicated interfaces for LLM-assisted engine scaffolding, where
generated drafts are reviewed, validated, and tested before being
integrated into workflows. At the same time, community governance will
be critical to ensure quality control, transparency, and reproducibility
of LLM-generated code. Beyond LLMs, the open architecture of the
framework makes it well-positioned for integration with new
computational backends, including GPU and cluster-based execution. By
combining distributed community involvement with emerging automation
tools, \pkg{flowengineR} is poised to evolve into a dynamic ecosystem
that adapts to both methodological and technological innovation.

\newpage



\noindent \textbf{Acknowledgement/Disclaimer:}

\par

\noindent This article is part of the cumulative doctoral thesis of
Maximilian Willer at the Carl von Ossietzky University Oldenburg. The
views expressed in this paper are those of the author and do not
necessarily reflect the views of HASPA Finanzholding. The research was
conducted independently and was not commissioned, funded, or reviewed by
HASPA Finanzholding.

\par\vspace{1em}

\noindent \textbf{Software used for this work and declaration of generative AI and AI-assisted technologies in the writing process:}

\par

\noindent The analyses and manuscript preparation were conducted using
RStudio v2025.09.1+401
(\citeproc{ref-positteamRStudioIntegratedDevelopment2025}{Posit team
2025}) with R v4.5.0 (\citeproc{ref-Rbase}{R Core Team 2025}) for
package development and Quarto-based document generation, GitHub
(\citeproc{ref-GitHub}{GitHub, Inc. 2025}) and GitLab
(\citeproc{ref-GitLab}{GitLab Inc. 2025}) for version control and
package management, Microsoft PowerPoint
(\citeproc{ref-MicrosoftPowerPoint}{Microsoft Corporation 2025}) for
figure preparation, and Chat Generative Pre-Trained Transformer
(\citeproc{ref-ChatGPT}{OpenAI 2025}) for text refinement, translation,
brainstorming, and literature exploration. The authors have carefully
reviewed and revised all formulations contained in this work and assume
full and sole responsibility for its content and statements.

\par\vspace{2em}



\section*{References}\label{bibliography}
\addcontentsline{toc}{section}{References}

\phantomsection\label{refs}

\end{document}